\documentclass[final,journal,twocolumn]{IEEEtran}
\ifCLASSOPTIONcompsoc
  \usepackage[nocompress]{cite}
\else
  \usepackage{cite}
\fi
\ifCLASSINFOpdf
   \usepackage[pdftex]{graphicx}
  \graphicspath{../figures/}
\else
\fi
\usepackage{amsmath}
\usepackage{upgreek}
\usepackage{amsfonts}
\usepackage{makecell}
\usepackage{array}
\usepackage{units}
\usepackage{multirow}

\usepackage{booktabs}
\usepackage{makecell}
\usepackage{tabularx}
\usepackage{url}

\usepackage{lipsum}
\usepackage{siunitx}
\usepackage{pifont}
\def\cmark{\ding{51}} 
\def\xmark{\ding{55}}

\newcommand\percentage[2][round-precision = 1]{
    \SI[round-mode = places,
        scientific-notation = fixed, fixed-exponent = 0,
        output-decimal-marker={.}, #1]{#2e2}{}
}

\newcommand\percentagex[2][round-precision = 2]{
    -
}

\newcommand\percentageBest[2][round-precision = 1]{
    {\bfseries \SI[round-mode = places,
        scientific-notation = fixed, fixed-exponent = 0, detect-inline-weight  = math, detect-weight = true,
        output-decimal-marker={.}, #1]{#2e2}{}}
}

\begin{document}
%

\title{SD-RSIC: Summarization Driven\\Deep Remote Sensing Image Captioning%
    \thanks{This work is funded by the European Research Council (ERC) through the ERC-2017-STG BigEarth Project under Grant 759764. \textit{(Corresponding author: Beg{\"u}m Demir).}}%
}
\author{%
    Gencer~Sumbul,~\IEEEmembership{Graduate Student Member,~IEEE},
    Sonali~Nayak,
	and~Beg{\"u}m~Demir~\IEEEmembership{Senior Member,~IEEE}%
    \thanks{Gencer Sumbul, Sonali Nayak and Beg{\"u}m Demir are with the Faculty of Electrical Engineering and Computer Science, Technische Universit\"at Berlin, 10623 Berlin, Germany.
    Email: \mbox{gencer.suembuel@tu-berlin.de}, \mbox{sonalirosy91@gmail.com}, \mbox{demir@tu-berlin.de}.}%
}

\maketitle

\begin{abstract}
Deep neural networks (DNNs) have been recently found popular for image captioning problems in remote sensing (RS). Existing DNN based approaches rely on the availability of a training set made up of a high number of RS images with their captions. However, captions of training images may contain redundant information (they can be repetitive or semantically similar to each other), resulting in information deficiency while learning a mapping from the image domain to the language domain. To overcome this limitation, in this paper, we present a novel Summarization Driven Remote Sensing Image Captioning (SD-RSIC) approach. The proposed approach consists of three main steps. The first step obtains the standard image captions by jointly exploiting convolutional neural networks (CNNs) with long short-term memory (LSTM) networks. The second step, unlike the existing RS image captioning methods, summarizes the ground-truth captions of each training image into a single caption by exploiting sequence to sequence neural networks and eliminates the redundancy present in the training set. The third step automatically defines the adaptive weights associated to each RS image to combine the standard captions with the summarized captions based on the semantic content of the image. This is achieved by a novel adaptive weighting strategy defined in the context of LSTM networks. Experimental results obtained on the RSCID, UCM-Captions and Sydney-Captions datasets show the effectiveness of the proposed approach compared to the state-of-the-art RS image captioning approaches. The code of the proposed approach is publicly available at \url{https://gitlab.tubit.tu-berlin.de/rsim/SD-RSIC}.
\end{abstract}


\section{Introduction}
\label{sec:introduction}
\IEEEPARstart{T}{he} new generation of remote sensing (RS) sensors characterized by very high geometrical resolution can acquire images with sub-metric spatial resolution. Thus, the significant amount of geometrical details can be presented in very high resolution RS image scenes. Accordingly, one of the most important applications is the RS image captioning, which aims at automatically assigning descriptive sentences (i.e., captions) to RS image scenes by accurately characterizing their semantic content. Recent studies in RS have shown that deep neural networks (DNNs) are capable of generating accurate image captions for RS images due to their ability to model a mapping from the high-level semantic content of RS images in image domain into the descriptive captions in language domain~\cite{Hoxha:2020}. DNN based encoder-decoder framework is one of the most effective methods for RS image captioning. Within this framework, image captioning is achieved based on two steps. In the first step, convolutional neural networks (CNNs) are used to extract image features, while in the second step recurrent neural network (RNN) based sequential approaches are used as a natural language model to generate a caption for each image based on the image features. The overall framework is considered as an encoder-decoder neural network where the encoder (CNN) takes an image as input and generates the corresponding encoded features, whereas the decoder generates a caption for the image based on the features. Then, the neural network trained on image-caption pairs can automatically generate a caption for a new image. Accordingly, in \cite{UCMCaptions}, CNNs and RNNs are employed to generate captions by combining image features of very high resolution RS images with the associated captions. In detail, pre-trained CNN models on a widely used computer vision dataset (i.e., ImageNet) are used to extract image features, while long-short term memory (LSTM) networks are utilized to sequentially characterize the image captions. In this study, two image captioning datasets are introduced as the first time in RS to evaluate the success of RS image captioning approaches. In \cite{traditionalrscap}, a conventional template-based method is presented in the context of RS image captioning for the cases where the number of RS images annotated with captions is not sufficient. This method represents RS images with a combination of ground elements, their attributes and relations that derive a language template. In detail, a fully convolutional network is introduced for the detection of multi-level ground elements, while captions are generated based on the predefined templates. In \cite{lu2017exploring}, the largest RS image captioning dataset, which is called RSICD, is introduced. In this study, traditional hand crafted features are compared with the features extracted through different CNN models in the context of RS image captioning, while the caption generation strategy introduced in \cite{UCMCaptions} is used. A Collective Semantic Metric Framework (CSMLF) that models the common semantic space of RS images and their captions is recently introduced in \cite{SemanticDescriptions}. In detail, CSMLF maps the GloVe based representations of image captions and the image features from a pre-trained CNN model into a common semantic space with a metric learning strategy. Then, the distance between a new image and all captions in the common space is computed to generate a new caption. In \cite{RSAttrAttn}, an attribute attention strategy that exploits the correlation between image regions and generated caption words is integrated into the standard encoder-decoder approach to further improve the semantic content characterization of images. In this approach, fully connected (FC) layers of a CNN are considered to characterize the image attributes, while convolutional layers are employed to obtain image features. The caption generation is achieved by using LSTMs (where the log likelihood of generating a caption word by word is maximized given the previous words), the image feature and corresponding image attributes. We would like note that although only a few DNN based RS image captioning approaches are proposed in RS literature, this research field has been extensively studied in computer vision. As an example, the above-mentioned encoder-decoder framework that jointly employs CNNs and RNNs for image captioning is initially introduced in \cite{GoogleNIC} as the first time. In \cite{showattendtell}, an attention mechanism is employed to characterize where or what to look in images to generate their captions. In \cite{Yu:2019}, topic embeddings are first extracted from a CNN-based multi-label classifier and then used with image features in an LSTM-based language model to generate topic-oriented image captions. We refer the readers to \cite{2019:Zakir} for a detailed review of DNN based image captioning approaches introduced in computer vision.

Most of the existing DNN based approaches in the context of RS image captioning rely on the availability of a training set, which consists of very high resolution RS images with their captions (which accurately describe the semantic content of images). Due to the complexity of learning in RS image and language domains, multiple captions are usually assigned to each training image to effectively and efficiently learn an image captioning model. Although each RS image is expected to be ideally described with different captions, each of which embodies different information of the image, a training set may contain redundant information through multiple captions. As an example, in the existing benchmark image captioning datasets (e.g., RSICD, Sydney-Captions and UCM-Captions), most of the RS images are associated with repetitive captions or similar captions with small differences. This can cause the information deficiency while learning a mapping from the image domain to the language domain. Redundant information in training sets may also lead to over-fitting in training, which reduces the generalization capability of image captioning models and thus causes poor image captioning performance. None of the existing DNN based approaches in RS take into account the above-mentioned problems. Thus, if a DNN model is trained on image caption pairs that include redundant information, existing captioning methods in RS may provide insufficient captioning performance. 

To overcome this limitation, in this paper, we introduce a novel Summarization Driven Remote Sensing Image Captioning (SD-RSIC) approach. The SD-RSIC aims at: i) learning to summarize image captions learned on large text corpora; and then ii) integrating it with the learning procedure of the captioning task to guide the whole training process. To this end, the proposed approach is made up of three main steps: 1) generation of standard captions; 2) summarization of ground-truth captions; and 3) integration of summarized captions with standard captions. In the first step, CNNs and LSTMs are jointly used as in the literature works for learning of standard image captions based on image features. In the second step, unlike the existing methods, we propose to exploit a sequence-to-sequence DNN model to summarize ground-truth captions of each image into a single caption. Due to this step, the proposed SD-RSIC approach is capable of eliminating redundant information present in captions, while enhancing the word vocabulary that provides more detailed captions for semantically complex RS images. In the third step, to integrate the summarized captions with the standard captions, the vocabulary word probabilities of standard captions are combined with those of the summarized captions based on the image features by a novel adaptive weighting strategy in the framework of LSTMs. This step reduces the risk of over-fitting during training, and thus improves the generalization capability of the whole approach.
The novelty of the proposed approach consists in: 1) summarization of ground-truth captions into single caption per RS image to eliminate the redundancy present in the ground-truth captions; 2) integration of the summarized captions with standard captions by an adaptive weighting strategy; and 3) exploiting the summarization approach that guides whole training procedure.

The rest of the paper is organized as follows: Section \ref{sec:method} provides the formulation of the image captioning task and introduces the proposed SD-RSIC approach. Section \ref{sec:data} describes the considered datasets, while Section \ref{sec:experiment} provides the experimental results. Section \ref{sec:conclusion} concludes our paper. 

\section{Proposed Summarization Driven Remote Sensing Image Captioning (SD-RSIC) Approach}\label{sec:method}

In this section, we first formulate the RS image captioning task, and then explain our Summarization Driven Remote Sensing Image Captioning (SD-RSIC) approach. Let $\mathcal{I} = \{I_1,\ldots,I_M\}$ be an archive that consists of $M$ images, where $I_i$ is the $i$\textsuperscript{th} image. We assume that a training set $\mathcal{T} \subset \mathcal{I}$ of images, each of which is annotated with one or more captions, is initially available. Let $C_i=\{c_{i,j}\}_{j=1}^{N_i}$ be the caption set associated with the $i$\textsuperscript{th} image $I_i$, where $c_{i,j}$ is the $j$\textsuperscript{th} caption of the set $C_i$ and $N_i$ is the number of considered captions. Each caption of the set $C_i$ can be formulated as the set of ordered words $c_{i,j}=\{w_{k}\}_{k=1}^{L_{i,j}}$, where $w_{k}$ is the $k$\textsuperscript{th} word in the caption and $L_{i,j}$ is the length of the caption $c_{i,j}$. The image captioning task aims to learn a function $F(I^*;\theta)$ that assigns a descriptive caption to a new image $I^*$. To this end, the parameters of the function can be learned by maximizing the log probability of the ground-truth captions for each $(I_i,C_i)$ training instance pair as follows:
\begin{equation} \label{eq1}
\theta^* = \operatorname*{arg\,max}_\theta \left(\sum_{i=1}^{|\mathcal{T}|} \sum_{j=1}^{N_i} \sum_{k=1}^{L_{i,j}} \log P(w_{k}|w_{1:k-1},I_i;\theta)\right)
\end{equation}
where $\theta$ is the whole parameter set of the function and $P(w_{k}|w_{1:k-1},I_i;\theta)$ is the probability of the $k$\textsuperscript{th} word $w_{k}$, which is conditioned on the previous words of the caption $c_{i,j}$ and the image ${I_i}$. Then, the caption of the image $I^*$ can be obtained by estimating the probabilities of corresponding words $P(w_k^*|w_{1:k-1}^*,I^*;\theta^*)$ with learned parameters. Conventional image captioning approaches in deep learning are based on encoder-decoder architectures for which the semantic content of RS images is encoded to facilitate the caption generation. 

\begin{figure*}[htb]
  \centering
  \includegraphics[width=0.99\linewidth]{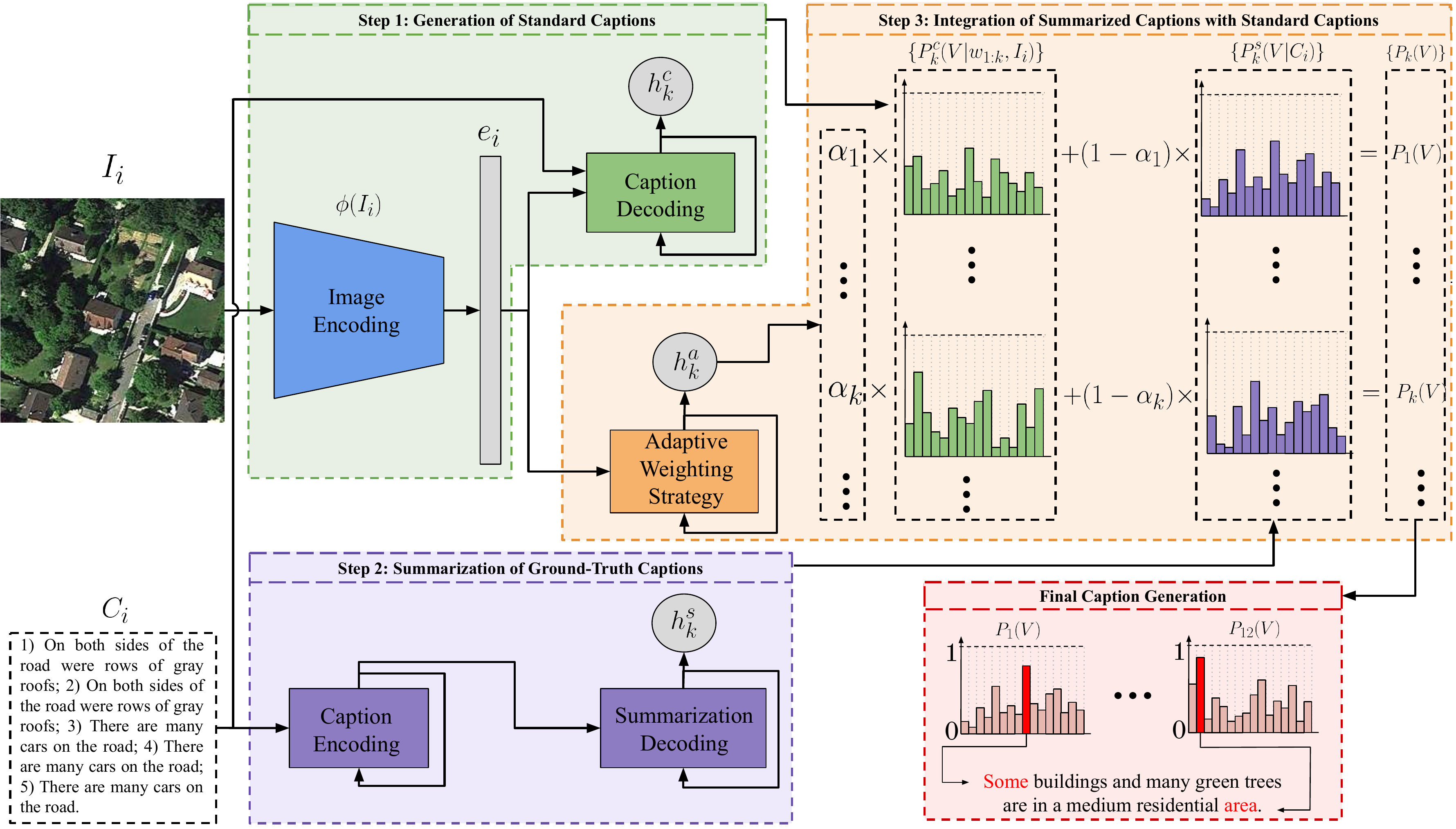}
  \caption{The proposed Summarization Driven Remote Sensing Image Captioning (SD-RSIC) approach.}
  \label{fig:model}
\end{figure*}
Learning image-caption mapping generally requires describing each image with many captions in the training set since by this way caption and image semantics can be accurately associated. However, the captions can share very similar semantics or include a large number of same words with similar orders. The disadvantages of redundant information present in ground-truth captions are twofold. First, this can cause the information deficiency during the learning process. Second, redundancy present in the captions can lead to over-fitting in training, which reduces the generalization capability of captioning models and thus causes poor image captioning performance. To address these problems, the proposed SD-RSIC approach is characterized by three main steps: 1) generation of standard captions; 2) summarization of ground-truth captions; and 3) integration of summarized captions with standard captions. The first step is based on the widely used learning method that jointly exploits CNNs and LSTMs for the image captioning problems. The novelty of the proposed SD-RSIC approach relies on the last two steps. In the second step, we propose to exploit sequence-to-sequence DNN models for the summarization of ground-truth image captions to eliminate the redundant information. In the third step, we introduce a novel adaptive weighting strategy to accurately define the weights for integrating the summarized captions with the standard captions according to the image  features. Fig. \ref{fig:model} presents a general overview of the proposed SD-RSIC approach and each step is explained in the following sections.

\subsection{Step 1: Generation of Standard Captions}
This step aims at generating consecutive words in a meaningful order that characterizes the standard image captions based on the image features. To this end, similar to the literature works in RS (e.g., \cite{lu2017exploring}), we utilize: i) CNNs to capture the high level semantic content of RS images; and ii) LSTMs to learn a mapping between the image features and consecutive word embeddings by sequentially modeling the language semantics. Let $\phi$ be any type of CNN. For a given image $I_i$, $\phi(I_i)$ provides a feature vector (i.e., image descriptor) to model the content of the image. In order to map the extracted feature vector to a common space with image captions, the extracted feature vector is given as input to a FC layer, which provides the final image embedding $e_i$ having the dimension of $W$. After the characterization of image features, an LSTM network produces a word at each time step based on the previous LSTM states and the word predictions to sequentially capture word semantics, while relying on the image features. At the beginning of the sequence, the image embedding $e_i$ is fed into the LSTM network that performs as the initial input of the sequence to affect the following word predictions. 
To start the caption sequence, we employ the special start token $w_0$ for all captions. Word generation is repeated until the special end token $w_e$ reaches to the network. To this end, we represent each word as a one-hot vector of dimension $|V|$, where $V$ is the 
\begin{figure}[t]
  \centering
  \includegraphics[width=1\columnwidth]{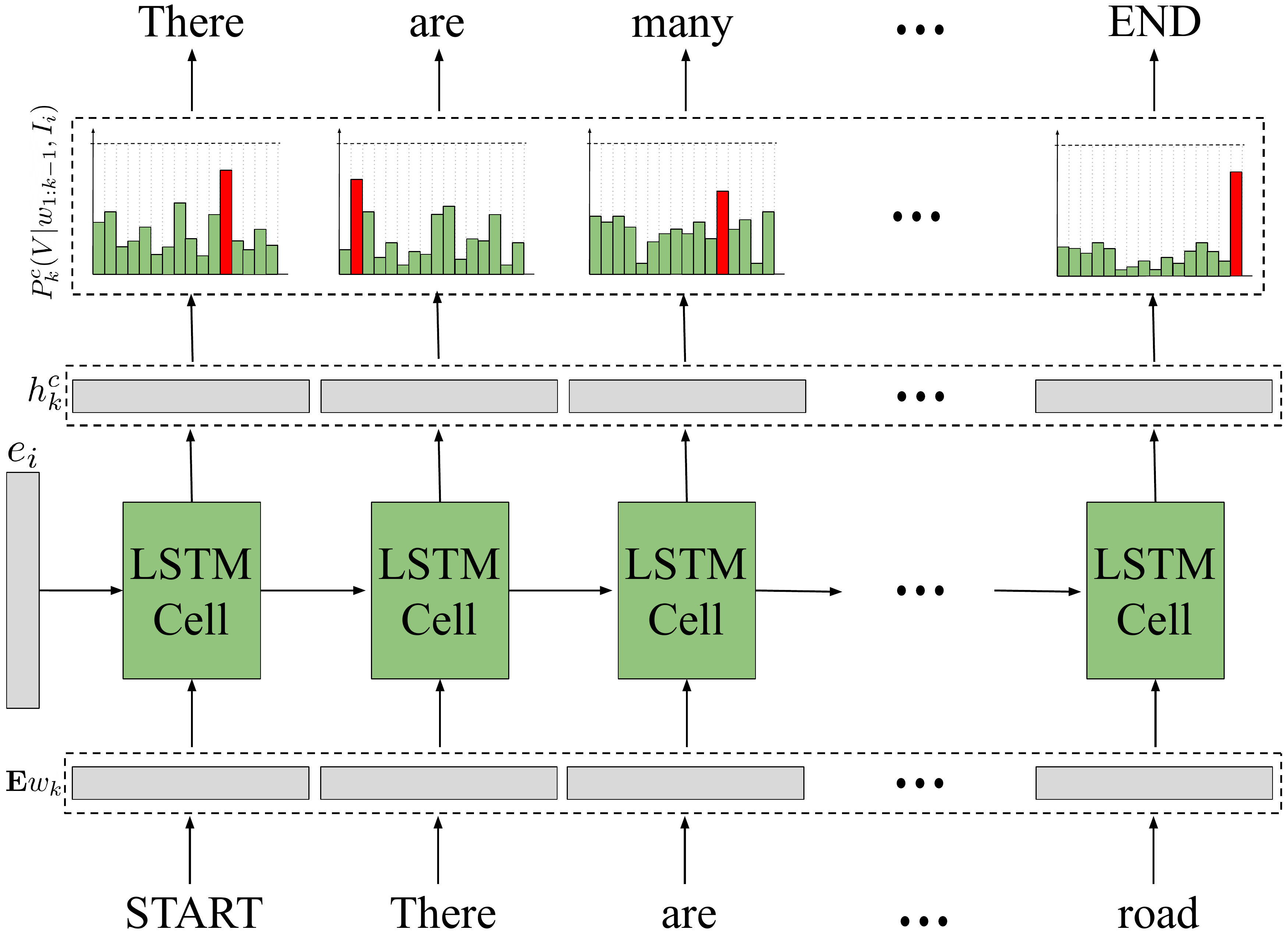}
  \caption{The first step of the SD-RSIC approach. The LSTM network used for this step is represented as unrolled, showing the input and output of a time step in the sequence.}
  \label{fig:captioning}
\end{figure}
vocabulary set including all unique words. In order to encode semantic similarity in words, we apply mapping from the one-hot vector representation into a real-valued embedding of words with the dimension of $W$ as follows: 
\begin{equation}
u_k=\mathbf{E}w_k, \quad w_k \in V
\end{equation}
where $\mathbf{E}$ is the word embedding matrix with the size of $W \times |V|$. The LSTM network of this step exploits word embedding and previous information of the sequence at each time step as follows:
\begin{equation}
  \begin{aligned}
    f_{k}  &=  \delta(\mathbf{W}_{f,u}u_k + \mathbf{U}_{f,h} h_{k-1}^c + b_{f})\\
    i_{k}  &=  \delta(\mathbf{W}_{i,u}u_k + \mathbf{U}_{i,h} h_{k-1}^c + b_{i})\\
    o_{k}  &=  \delta(\mathbf{W}_{o,u}u_k + \mathbf{U}_{o,h} h_{k-1}^c + b_{o})\\
    c_{k}^c  &=  f_{k}\odot c_{k-1}^c  + i_{k}\odot\tanh(\mathbf{W}_{c,u}u_k \\ &\phantom{{}=f_{k}\odot c_{k-1}} + \mathbf{U}_{c,h} h_{k-1}^c + b_{c})\\
    h_{k}^c  &=  o_{k} \odot \tanh(c_{k}^c) \\
  \end{aligned}
\end{equation}
where $\mathbf{W}_{\boldsymbol{.}}$ and $b_{\boldsymbol{.}}$ are the weight and bias parameters, respectively. $\tanh$ and $\delta$ are the hyperbolic tangent and sigmoid functions, and $i$, $f$, $o$ and $c$ are input gate, forget gate, output gate and cell state, respectively (for a detailed explanation, see~\cite{lstm, Gers:2000}). At the beginning of the sequence, $c_0^c$ and $h_0^c$ are randomly initialized. Then, we obtain word probabilities at each time step with softmax function following to a classification layer as follows:
\begin{equation}
P_k^c(V|w_{1:k-1},I_i;\theta)  =  \sigma(\mathbf{W}_{p,h}h_{k}^c + b_p)
\end{equation}
where $\sigma$ is the softmax function and $\mathbf{W}_{p,h}$ and ${b_p}$ are the weight and bias parameters of a FC layer. $P_k^c(V|w_{1:k-1},I_i;\theta)$ denotes the probability distribution of all vocabulary words produced at the $k$\textsuperscript{th} time step of the corresponding LSTM network. This step is illustrated in Fig. \ref{fig:captioning}.

\subsection{Step 2: Summarization of Ground-Truth Captions}
This step aims to summarize the ground-truth captions of RS images. The summarized captions guide the whole training process of the proposed SD-RSIC approach. To this end, we propose to adapt the automatic summarization task of natural language processing literature into the image captioning problem. The summarization task is defined as condensing a text to a shorter version that contains the most important information. In our approach, we exploit pointer-generator DNNs \cite{PointerGenerator17} as a special type of sequence-to-sequence neural networks. To this end, we consider to train the pointer-generator model on news articles to automatically extract headlines. Then, we exploit the model for summarizing ground-truth captions in our approach. To this end, we stack all corresponding  captions of each RS image as a single text to summarize them into single caption. Then, all words of stacked captions are embedded as in (2) and fed into the pre-trained model. Two recurrent neural networks sequentially encode the stacked captions and decode them to generate a summarized caption in order. In addition, pointer-generator structure decides the probability of generating words from the vocabulary versus copying from all captions. This allows an accurate reproduction of information, while retaining the ability to produce novel words through the generator (for a detailed explanation, see~\cite{PointerGenerator17}). Let $\psi$ be the pre-trained summarization network, $\psi(\{c_{i,j}\}_{j=1}^{N_i})$ produces the word probabilities of the vocabulary $P_k^s(V|\{c_{i,j}\}_{j=1}^{N_i})$ at the $k$\textsuperscript{th} time step. 

Due to the summarization of ground-truth captions, the proposed SD-RSIC approach is capable of eliminating redundant information present in the multiple captions associated with each training image by condensing all captions into a single caption that captures the most significant information content. In addition, the summarization model is pre-trained on a dataset whose vocabulary is excessively larger than any RS image captioning dataset. In this way, our approach uses significantly bigger vocabulary (which is also used in all steps of the SD-RSIC) compared to existing approaches. Using enriched vocabulary increases the capability of our approach to generate more detailed captions for semantically complex RS images.

\subsection{Step 3: Integration of Summarized Captions with Standard Captions}
This step aims to define a final caption for each image by reducing the limitations of redundant information in ground-truth captions, while providing the detailed language semantics. To this end, we propose to integrate the standard caption of each image with its summarized caption based on a novel adaptive weighting strategy. The proposed strategy employs an LSTM network, which automatically characterizes the weights for combining the vocabulary word probabilities of standard captions with those of the summarized captions at each time step. Initial cell state $c^a_0$ and hidden state $h^a_0$ of the LSTM network are randomly initialized, and then the LSTM takes the final image embedding $e_i$ as input at each time step. Then, a single weight score $h^a_t$ is produced as in (3) at each time step based on the previous cell states and the image embedding. To normalize the scores to the range of $[0,1]$, we apply sigmoid function to obtain the final weights $\{\alpha_k\}_{k=1}^{N_i}$ for the RS image $I_i$. Then, final word probability distribution at time step $k$ is obtained by the weighted combination of the word probabilities of standard captions (which is obtained in the first step) and those obtained in the second step as follows:
\begin{equation}
P_k(V) = \alpha_k \times P_k^c(V|w_{1:k-1},I_i) + (1 - \alpha_k) \times  P_k^s(V|C_i).
\end{equation}  
If there is no corresponding output in the first or second step at the $k$\textsuperscript{th} time step, we apply zero-padding to the shorter output. After obtaining the probabilities for all time steps, we achieve the final caption by selecting the words leading to the highest probabilities. Since the learning of the weights is achieved based on the image features, weights are adaptive depending on the content of the images, i.e., different weights are assigned to different images. Due to the proposed adaptive weighting strategy, the proposed SD-RSIC approach is capable of exploiting the summarized captions to guide the training of the whole neural network. With this guidance, the training procedure is less affected by the redundancy present in the ground-truth captions. This process: i) reduces the risk of over-fitting and thus increases the generalization capability of the SD-RSIC; and ii) thus leads to a more effective learning procedure and more accurate RS image captions.

For the training of the proposed SD-RSIC approach, we use the stochastic gradient descent based optimization to maximize the log probability of the ground-truth captions for each $(I_i,C_i)$ training instance using (1). After learning model parameters, the proposed approach automatically generates a caption for a new RS image. This process does not require any ground-truth caption since the summarization of ground-truth captions is only applied in the training stage. It is worth noting that finding the optimal word sequence is computationally expensive during the inference due to a large number of possible output sequences. Thus, we utilize the beam search algorithm with a beam size of four to acquire the best word sequence. This algorithm iteratively considers the set of best captions up to $k$\textsuperscript{th} time step to produce the captions for the time step of $k+1$. However, it keeps only some of them depending on the beam size parameter value. 

We would like to note that the summarized and standard captions can be semantically different (mainly due to possible differences between the lengths of the captions). However, since the adaptive weights of the words are iteratively learned, the proposed approach is not significantly affected by the possible semantic differences between the summarized and the standard captions. In detail, when the optimization process converges, the weights become more adapted to compensate the semantic differences between the summarized and the standard captions. In addition, iteratively learning the weights also forces generated weights and the standard captions to be in the same semantic order.
\section{Datasets and Experimental Setup}\label{sec:data}
In this section, we first describe the datasets used in the experiments and then present the experimental setup with the description of the baseline approaches. 
\begin{figure}[t]
  \centering
  \includegraphics[width=1\columnwidth]{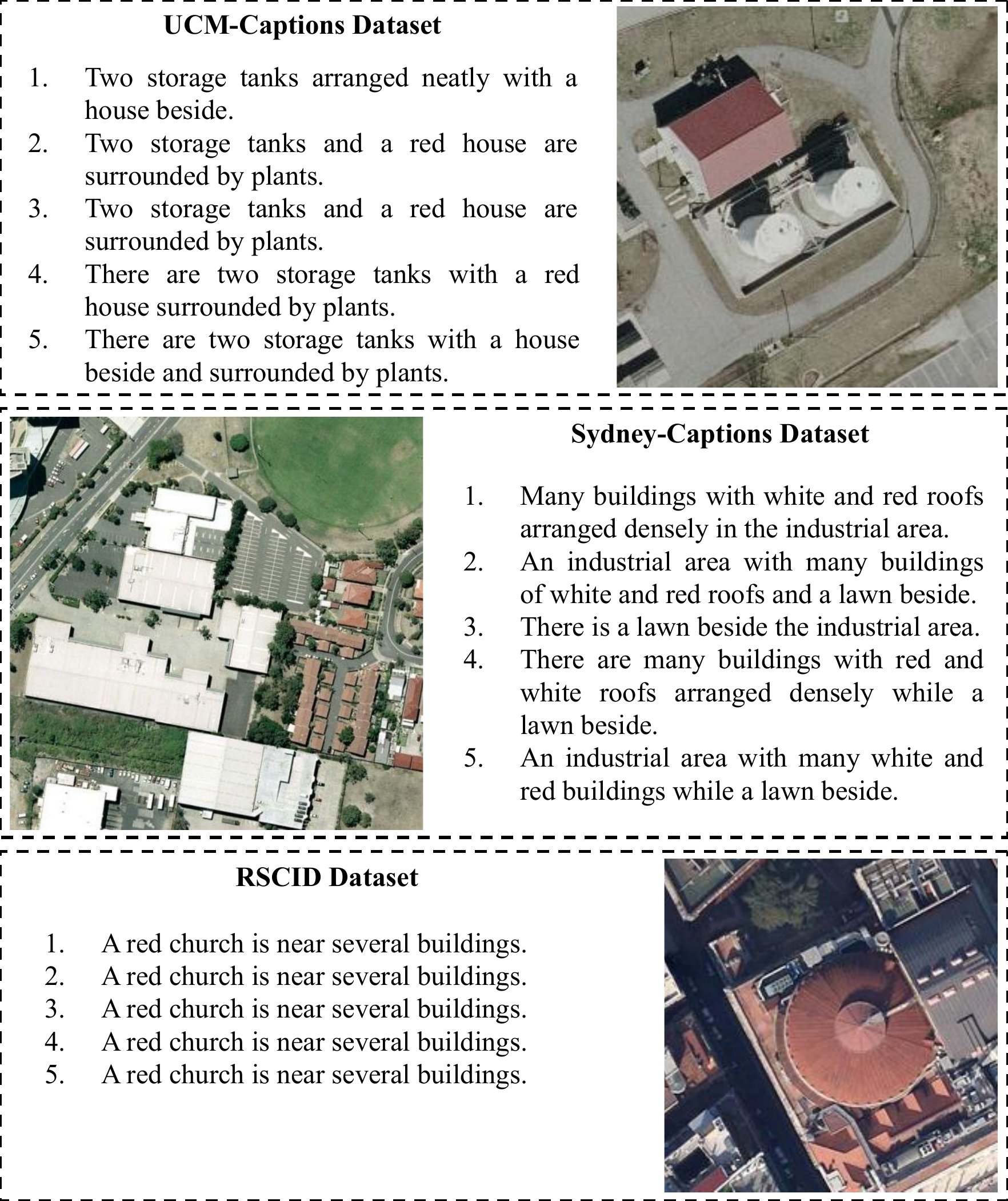}
  \caption{An example of RS images with their ground-truth captions selected from the UCM-Captions (top), the Sydney-Captions (middle) and the RSICD (bottom) datasets.}
  \label{fig:dataset}
\end{figure}
\subsection{Dataset Description}
\begin{table}
\renewcommand{\arraystretch}{0.3}
\centering
\caption{An Example of Article-Headline Pairs in the Annotated Gigaword dataset}
\label{table:gigaword}
\begin{tabular}{p{5.6cm}p{2.3cm}}
\midrule 
{\centering \textbf{Article}} & {\centering \textbf{Headline}} \\
\midrule
A fire on a freight shuttle in the channel tunnel on thursday forced an emergency rescue operation and the closure of the tunnel, officials said. & Fire closes channel tunnel \\
\midrule
World oil prices rose in asian trade thursday as hurricane ike headed towards key energy facilities on the southern us coast, dealers said. &  Oil prices up in asia on hurricane fears \\
\midrule
\end{tabular}
\end{table}
To evaluate our approach, we performed experiments on the Sydney-Captions\cite{UCMCaptions}, UCM-Captions\cite{UCMCaptions} and RSICD\cite{lu2017exploring} datasets. In addition, we utilized the Annotated Gigaword dataset \cite{gigaword}, \cite{Napoles} for the second step of the proposed SD-RSIC approach.

The Sydney-Captions dataset includes 613 images, each of which has the size of 500$\times$500 pixels with a spatial resolution of 0.5 meters. This dataset was built based on the Sydney scene classification dataset~\cite{sydneydataset}, which includes RS images annotated with one of the seven land-use classes. Each image in the Sydney-Captions dataset was annotated by the five captions, providing 3065 captions in total. The UCM-Captions dataset includes 2100 aerial images, each of which has a size of 256$\times$256 pixels with a spatial resolution of one foot. This dataset is defined based on the UC Merced Land Use dataset \cite{ucmdataset}, in which each image is associated with one of 21 land-use classes. Each image in the UCM-Captions dataset was annotated with five captions, resulting in 10500 captions in total. Although five captions per image are considered, captions belonging to the same classes are very similar in both datasets. Both the Sydney-Captions and the UCM-Captions datasets were initially built for scene classification problems with a small number of images. The RSICD is currently the largest RS image captioning dataset, including 10921 images in total with the size of 224$\times$224 pixels with varying spatial resolutions. In this dataset, each image is described with a different number of captions \cite{lu2017exploring}. In detail, 724 images have five different captions, 1495 images have four different captions, 2182 images have three different captions, 1667 images have two different captions and 5853 images have only one caption. As mentioned in \cite{lu2017exploring}, the number of captions was augmented in cases where images are described with less than five captions by randomly duplicating the existing captions. This leads to 54605 captions in the dataset. Fig. \ref{fig:dataset} shows an example of images and their captions for all considered RS image captioning datasets. The Annotated Gigaword dataset is a corpus of article-headline pairs that consists of nearly 10 million documents with a total of more than 4 billion words sourced from various news services. Instead of using the whole corpus, we follow the same removal and pre-processing steps presented in \cite{Rush:2015} that results in around 4 million articles. Table \ref{table:gigaword} shows an example of article-headline pairs in this dataset.

\subsection{Experimental Setup}
To perform the experiments, we split each considered dataset into training (80\%), validation (10\%) and test (10\%) sets as suggested in the papers that the datasets were introduced (\!\!~\cite{UCMCaptions, lu2017exploring}). All hyper-parameters were obtained based on the RS image captioning performance on the validation set. In the training sets of all datasets, there are five captions per image. Thus, we replicated each image five times to compose image-caption pairs of training. For the Annotated Gigaword dataset, we initially used the same training set splitting with \cite{Rush:2015} that results in 110,000 unique words, which is significantly higher than any vocabulary size within the RS captioning datasets. Then, we changed the vocabulary set of captioning datasets, since they do not contain all the words from the summarization vocabulary, and thus might miss several words when we summarize the five captions to one using the summarization model. Accordingly, we constructed a new common vocabulary set, which is used in all the steps of our approach. To this end, we selected 50000 words that include all the words from the Sydney-Captions, UCM-Captions and RSICD datasets and the list of most appearing words in the Annotated Gigaword dataset.

Before training our approach, we trained the pointer-generator network for summarization by following the same hyper-parameters presented in \cite{PointerGenerator17}. Then, we combine the pre-trained model with our approach. In addition, we also utilized the existing CNN models, which are pre-trained on the ImageNet for the feature extractor $\phi$ in the first step of the SD-RSIC. To select the CNN model for each dataset, $\phi$ is tested among the CNNs of the VGG~\cite{Simonyan:2015}, GoogleNet~\cite{Szegedy:2015}, InceptionV3~\cite{Szegedy:2016}, ResNet~\cite{He:2016} and DenseNet~\cite{Huang:2017} models. We would like to note that we did not apply fine-tuning to the parameters of pre-trained models during the training of our approach. The extracted image features are mapped to the embedding space, whose dimension is the same as the word embedding dimension. In the experiments, the value of the embedding size $W$ is varied as $W$ = 128, 256, 512, 1024. However, for the selection of $\phi$, the value of the embedding size is fixed to 512. In the first and third steps of our approach, we exploited the LSTM networks with $W$ and 1 dimensional hidden states, respectively. We trained our approach with the learning of $10^{-3}$, which decays by $20\%$ if there are eight consecutive epochs without any improvement on the validation set performance. The training was conducted on NVIDIA Tesla V100 GPUs. To assess the effectiveness of the second and the third steps of the proposed approach, we considered a scenario for which these steps are neglected and only the first step of the proposed approach is applied. For this scenario, we randomly selected a single caption for each image in the training sets of all the considered datasets. It is denoted as Step 1 (Single Caption) in the experiments. To assess the effectiveness of the different steps of the proposed approach in terms of computational complexity, we provided the total number of parameters and floating-point operations associated to the different steps of the proposed approach.

In the experiments, we compared our approach with: 1) the cosine distance matching between the bag-of-words representation of image captions and the CNN features of images (which is denoted as BoW+CNN); 2) the cosine distance matching between the Deep Visual-Semantic Embedding (DeViSE)~\cite{Frome:2013} of image captions and the CNN features of images (which is denoted as DeViSE+CNN); 3) the Collective Semantic Metric Learning Framework (CSMLF)~\cite{SemanticDescriptions}; and 4) the Neural Image Caption (NIC) \cite{GoogleNIC}. RS image captioning accuracies of the BoW+CNN, DeViSE+CNN and CSMLF on each dataset were obtained in~\cite{SemanticDescriptions} by utilizing the ResNet model at the depth of 50 (ResNet50) as the feature extractor for RS images. Since the results were obtained by using the same sets with our approach, we did not repeat the corresponding experiments. For the NIC, which is one of the widely used state-of-the-art RS image captioning approaches, we applied the same CNN and caption generation procedure as the first step of our approach for each experiment of the NIC to fairly compare it with the proposed SD-RSIC approach.

\begin{table*}[t]
\renewcommand{\arraystretch}{0.7}
\centering
\caption{Image Captioning Performance on the Sydney-Captions Dataset When Using Different CNN Models for the Proposed SD-RSIC Approach}
\label{table:sens_sydney}
\begin{tabular}{lccccccc}
\cmidrule[.8pt]{1-8}
Method & {\centering BLEU-1}
& {\centering BLEU-2}
& {\centering BLEU-3}
& {\centering BLEU-4}
& {\centering METEOR}
& {\centering ROUGE-L}
& {\centering CIDEr}
\tabularnewline
\cmidrule[.8pt]{1-8}
VGG16 & \percentage{0.724} & \percentage{0.621} & \percentage{0.532} & \percentage{0.451} & \percentage{0.342} & \percentage{0.636} & \percentage{1.395}
\tabularnewline\midrule
VGG19 & \percentage{0.734} & \percentage{0.631} & \percentage{0.552} & \percentage{0.487} & \percentage{0.348} & \percentage{0.641} & \percentage{1.603}
\tabularnewline\midrule 
GoogleNet & \percentage{0.715} & \percentage{0.605} & \percentage{0.511} & \percentage{0.422} & \percentage{0.333} & \percentage{0.628} & \percentage{1.306}
\tabularnewline\midrule 
InceptionV3 & \percentage{0.733} & \percentage{0.626} & \percentage{0.545} & \percentage{0.477} & \percentage{0.351} & \percentage{0.629} & \percentage{1.439}
\tabularnewline\midrule 
ResNet34 & \percentage{0.730}  & \percentage{0.629}  & \percentage{0.544}  & \percentage{0.468}  & \percentage{0.343}  & \percentage{0.637}  & \percentage{1.376}
\tabularnewline\midrule
ResNet50 & \percentage{0.716} & \percentage{0.592} & \percentage{0.491} & \percentage{0.398} & \percentage{0.320} & \percentage{0.616} & \percentage{1.087}
\tabularnewline\midrule
ResNet101 & \percentage{0.761} & \percentage{0.666} & \percentage{0.586} & \percentage{0.517} & \percentage{0.366} & \percentage{0.657} & \percentage{1.690}
\tabularnewline\midrule
ResNet152 & \percentage{0.733} & \percentage{0.619} & \percentage{0.517} & \percentage{0.425} & \percentage{0.318} & \percentage{0.620} & \percentage{1.146}
\tabularnewline\midrule
DenseNet121 & \percentage{0.736} & \percentage{0.634} & \percentage{0.552} & \percentage{0.478} & \percentage{0.349} & \percentage{0.638} & \percentage{1.389}
\tabularnewline\midrule
DenseNet169 & \percentage{0.730} & \percentage{0.632} & \percentage{0.546} & \percentage{0.467} & \percentage{0.341} & \percentage{0.629} & \percentage{1.402}
\tabularnewline\midrule
DenseNet201 & \percentage{0.718} & \percentage{0.616} & \percentage{0.532} & \percentage{0.453} & \percentage{0.333} & \percentage{0.624} & \percentage{1.378}
\tabularnewline
\cmidrule[.8pt]{1-8}
\end{tabular}
\end{table*}

Results of each experiment are provided in terms of four performance evaluation metrics: 1) the Bilingual Evaluation Understudy (BLEU) \cite{BLEU}, 2) the Meteor Universal (METEOR) \cite{Meteor2014}, 3) the Longest Common Subsequence-Based F-Measure of Recall-Oriented Understudy for Gisting Evaluation (ROUGE-L) \cite{Rouge2004} and 4) the Consensus-Based Image Description Evaluation (CIDEr) \cite{CIDEr2014}.

BLEU is not only the oldest but also the most well-known metric used for sentence similarity measurement. It measures the closeness of machine translation with one or more reference human translation according to numerical metrics that is proposed in~\cite{BLEU}. It compares $n$-grams of machine generated captions with the $n$-grams of ground-truth captions and then counts the number of matches. Thus, the score is better if machine translation is closer to human translation. It is calculated by finding the geometric mean of $n$-gram precision scores as follows:
\begin{equation}
    \text{BLEU-}n =  \text{BP} \times e^{( \sum_{n=1}^{N^B} w_n^B \log P_n^B)}
\end{equation}
where $P_n^B$ and $w_n^B$ are the precision and weights of $n$-grams. It further applies brevity penalty $\text{BP}$ for short sentences as follows:
\begin{equation}
    \text{BP} =  \Bigg\{ \begin{array}{c}
    1 \quad\quad\quad\quad\ \ if \quad l_c > l_r\\
    e^{(1-l_r/l_c)} \quad if \quad l_c \leq l_r\\
    \end{array}
\end{equation}
where $l_c$ and $l_r$ are the lengths of the candidate and ground-truth captions, respectively.

METEOR is based on word-to-word matching scores. For the multiple ground-truth captions, the score is calculated with respect to each caption and the best score is considered only. First, an $F$-Score ($F^M$) is calculated based on the word-to-word matching precision ($P^M$) and recall ($R^M$) scores as follows:
\begin{equation}
    F^M = \frac{10 \times P^M \times R^M}{R^M+9\times P^M}.
\end{equation}
Then, METEOR is calculated as follows:
\begin{equation}
    \text{METEOR} = F^M \times (1 - \frac{0.5 \times |\text{Chunks}|}{|\text{Matched Words}|})
\end{equation}
where chunk is defined as a series of contiguous and identically ordered matches among the candidate and ground-truth captions.

ROUGE-L considers the longest common sub-sequence (LCS) between a pair of candidate and ground-truth captions. It is a type of $F$-Score based on the precision ($P^L$) and recall ($R^L$) scores of LCS results as follows:
\begin{equation}
  \begin{aligned}
R^L &= \frac{|LCS|}{l_r} \\
P^L &= \frac{|LCS|}{l_c} \\
\text{ROUGE-L} &= \frac{(1 + \beta^2) \times R^L \times P^L}{R^L + \beta^2 \times P^L}.\\
\end{aligned}
\end{equation}

CIDEr considers a consensus of how often the $n$-grams in a candidate caption is present in ground-truth captions. It also considers the $n$-grams, which are not present in the ground-truth captions and should not be presented in the candidate caption~\cite{CIDEr2014}. To this end, it is calculated based on the Term Frequency Inverse Document Frequency (TF-IDF) weighting for each $n$-gram as follows:
\begin{equation}
 \begin{aligned}
    \text{CIDEr}_n &= \frac{1}{m} \sum_{j} \frac{g^n(c_i^*) \cdot g^n(c_{i,j})}{||g^n(c_i^*)|| \: ||g^n(c_{i,j})||}\\
	\text{CIDEr} &= \sum_{n=1}^N w_n^B \text{CIDEr}_n\\
\end{aligned}
\end{equation}
where $c_i^*$ and $c_{i,j}$ are the candidate and ground-truth captions, respectively and $g_n$ is a function that provides the vector of all $n$-grams of length $n$.

\section{Experimental Results}
\label{sec:experiment}
\begin{table*}[t]
\renewcommand{\arraystretch}{0.7}
\centering
\caption{Image Captioning Performance on the UCM-Captions Dataset When Using Different CNN Models for the Proposed SD-RSIC Approach}
\label{table:sens_ucm}
\begin{tabular}{lccccccc}
\cmidrule[.8pt]{1-8}
Method & {\centering BLEU-1}
& {\centering BLEU-2}
& {\centering BLEU-3}
& {\centering BLEU-4}
& {\centering METEOR}
& {\centering ROUGE-L}
& {\centering CIDEr}
\tabularnewline
\cmidrule[.8pt]{1-8}
VGG16 & \percentage{0.748} & \percentage{0.664} & \percentage{0.598} & \percentage{0.538} & \percentage{0.390} & \percentage{0.695} & \percentage{2.132}
\tabularnewline\midrule
VGG19 & \percentage{0.734} & \percentage{0.652} & \percentage{0.583} & \percentage{0.522} & \percentage{0.370} & \percentage{0.679} & \percentage{2.080}
\tabularnewline\midrule 
GoogleNet & \percentage{0.746} & \percentage{0.653} & \percentage{0.583} & \percentage{0.526} & \percentage{0.378} & \percentage{0.675} & \percentage{2.149}
\tabularnewline\midrule 
InceptionV3 & \percentage{0.694} & \percentage{0.591} & \percentage{0.516} & \percentage{0.456} & \percentage{0.334} & \percentage{0.622} & \percentage{1.734}
\tabularnewline\midrule 
ResNet34 & \percentage{0.733}  & \percentage{0.636}  & \percentage{0.560}  & \percentage{0.496}  & \percentage{0.362}  & \percentage{0.662}  & \percentage{1.971}
\tabularnewline\midrule
ResNet50 & \percentage{0.743} & \percentage{0.654} & \percentage{0.582} & \percentage{0.515} & \percentage{0.358} & \percentage{0.667} & \percentage{2.057}
\tabularnewline\midrule
ResNet101 & \percentage{0.722} & \percentage{0.633} & \percentage{0.561} & \percentage{0.499} & \percentage{0.363} & \percentage{0.667} & \percentage{1.998}
\tabularnewline\midrule
ResNet152 & \percentage{0.714} & \percentage{0.625} & \percentage{0.553} & \percentage{0.492} & \percentage{0.363} & \percentage{0.658} & \percentage{1.978}
\tabularnewline\midrule
DenseNet121 & \percentage{0.726}  & \percentage{0.631}  & \percentage{0.556}  & \percentage{0.491}  & \percentage{0.357}  & \percentage{0.658}  & \percentage{1.967}
\tabularnewline\midrule
DenseNet169 & \percentage{0.747} & \percentage{0.653} & \percentage{0.581} & \percentage{0.518} & \percentage{0.375} & \percentage{0.681} & \percentage{2.028}
\tabularnewline\midrule
DenseNet201 & \percentage{0.731} & \percentage{0.635} & \percentage{0.562} & \percentage{0.498} & \percentage{0.353} & \percentage{0.653} & \percentage{1.955}
\tabularnewline
\cmidrule[.8pt]{1-8}
\end{tabular}
\end{table*}
\begin{table*}[t]
\renewcommand{\arraystretch}{0.7}
\centering
\caption{Image Captioning Performance on the RSICD Dataset When Using Different CNN Models for the Proposed SD-RSIC Approach}
\label{table:sens_RSICD}
\begin{tabular}{lccccccc}
\cmidrule[.8pt]{1-8}
Method & {\centering BLEU-1}
& {\centering BLEU-2}
& {\centering BLEU-3}
& {\centering BLEU-4}
& {\centering METEOR}
& {\centering ROUGE-L}
& {\centering CIDEr}
\tabularnewline
\cmidrule[.8pt]{1-8}
VGG16 & \percentage{0.645} & \percentage{0.471} & \percentage{0.364} & \percentage{0.294} & \percentage{0.249} & \percentage{0.519} & \percentage{0.775}    \tabularnewline\midrule
VGG19 & \percentage{0.648} & \percentage{0.473} & \percentage{0.365} & \percentage{0.293} & \percentage{0.251} & \percentage{0.518} & \percentage{0.765}  \tabularnewline\midrule 
GoogleNet & \percentage{0.637} & \percentage{0.457} & \percentage{0.349} & \percentage{0.280} & \percentage{0.244} & \percentage{0.510} & \percentage{0.736} \tabularnewline\midrule 
InceptionV3 & \percentage{0.628} & \percentage{0.450} & \percentage{0.343} & \percentage{0.273} & \percentage{0.238} & \percentage{0.507} & \percentage{0.718} \tabularnewline\midrule
ResNet34 & \percentage{0.635}  & \percentage{0.460}  & \percentage{0.352}  & \percentage{0.282}  & \percentage{0.242}  & \percentage{0.511}  & \percentage{0.738} \tabularnewline\midrule
ResNet50 & \percentage{0.649} & \percentage{0.472} & \percentage{0.365} & \percentage{0.295} & \percentage{0.249} & \percentage{0.520} & \percentage{0.773} \tabularnewline\midrule
ResNet101 & \percentage{0.631} & \percentage{0.458} & \percentage{0.354} & \percentage{0.287} & \percentage{0.241} & \percentage{0.512} & \percentage{0.754}  \tabularnewline\midrule
ResNet152 & \percentage{0.644} & \percentage{0.474} & \percentage{0.369} & \percentage{0.300} & \percentage{0.249} & \percentage{0.523} & \percentage{0.794} \tabularnewline\midrule
DenseNet121 & \percentage{0.632}  & \percentage{0.462}  & \percentage{0.358}  & \percentage{0.288}  & \percentage{0.245}  & \percentage{0.514}  & \percentage{0.757} \tabularnewline\midrule
DenseNet169 & \percentage{0.643} & \percentage{0.465} & \percentage{0.357} & \percentage{0.285} & \percentage{0.244} & \percentage{0.512} & \percentage{0.759}  \tabularnewline\midrule
DenseNet201 & \percentage{0.625} & \percentage{0.457} & \percentage{0.351} & \percentage{0.281} & \percentage{0.240} & \percentage{0.513} & \percentage{0.742} \tabularnewline
\cmidrule[.8pt]{1-8}
\end{tabular}
\end{table*}

We carried out different kinds of experiments in order to: 1) perform a sensitivity analysis; and 2) compare the effectiveness of the proposed SD-RSIC approach with the state-of-the-art image captioning approaches. 

\subsection{Sensitivity Analysis of the Proposed Approach}
In this sub-section, we perform the sensitivity analysis of the proposed SD-RSIC approach in terms of: i) different CNN models utilized in the first step; ii) different embedding size used for image features and captions; iii) the effectiveness of the second and third steps; iv) the computational complexity associated to the different steps; and v) the sensitivity to zero-padding operation applied in the third step.

In the first set of trials, we analyzed the effect of different CNN models (the VGG model at the depths of 16 and 19 layers [VGG16, VGG19], the GoogleNet model, the InceptionV3 model, the ResNet model at the depths of 34, 50, 101 and 152 layers [ResNet34, ResNet50, ResNet101, ResNet152] and the DenseNet model at the depths of 121, 169 and 201 layers [DenseNet121, DenseNet169, DenseNet201]) in the first step of the proposed approach in terms of the image captioning performance. Table \ref{table:sens_sydney} shows the results for the Sydney-Captions dataset. By assessing the table, one can observe that the ResNet model at the depth of 101 layers leads to the highest scores under all metrics compared to the other CNNs. As an example, the ResNet101 provides almost $5\%$ higher BLEU-1, more than $6\%$ higher BLEU-2, almost $8\%$ higher BLEU-3, more than $9\%$ higher BLEU-4 and almost $3\%$ higher ROUGE-L scores compared to the GoogleNet model. In detail, most of the CNN models (except ResNet101) achieve similar scores on the Sydney-Captions dataset under all metrics regardless of their depth. As an example, the VGG model at the lowest depth in considered CNNs (VGG16) provides less than $1\%$ higher BLEU-1 and almost the same BLEU-4 scores compared to the DenseNet model at the highest depth among all CNNs (DenseNet201). The image captioning results for the UCM-Captions dataset is given in Table \ref{table:sens_ucm}. By analyzing the table, one can see that the VGG model at the depth of 16 layers (VGG16) provides the highest scores under all metrics except CIDEr. As an example, the VGG16 provides more than $5\%$ higher BLEU-1, more than $8\%$ higher BLEU-4 and more than $7\%$ higher ROUGE-L scores compared to the InceptionV3. However, only under CIDEr metric, the VGG16 leads to less than $2\%$ lower score compared to the highest score obtained by the GoogleNet model. In detail, the InceptionV3  provides the lowest scores under all metrics. As an example, it provides more than $5\%$ lower BLEU-1 and almost $6$ lower ROUGE-L scores compared to the DenseNet169. These results show that almost all CNN models (except the InceptionV3) achieve similar scores on the UCM-Captions dataset. This supports our conclusion on the Sydney-Captions dataset. In greater details, increasing the depths of the ResNet and DenseNet models up to some extent achieves slightly higher metric scores compared to those at the lowest depth. However, further increasing their depths do not provide the highest scores. As an example, the ResNet model at the depth of 152 leads to the lowest score under most of the metrics compared to the other ResNet CNNs. Table \ref{table:sens_RSICD} shows the results for the RSICD dataset. By analyzing the table, one can observe that the ResNet model at the depth of 152 layers provides the highest scores under most of the metrics compared to the other CNNs. As an example, the ResNet152 achieves more than $2\%$ higher BLEU-3 and BLUE-4 scores and almost $8\%$ higher CIDEr score compared to the InceptionV3. It also achieves almost the same BLEU-1 and METEOR scores with the VGG19 and the ResNet50, which provide the highest score in BLEU-1 and METEOR metrics, respectively. In detail, the VGG model (which has the shallowest CNNs compared to the others) leads to higher scores under most of the metrics compared to the DenseNet model. As an example, the VGG model at the depth of 19 layers achieves more than $2\%$ BLEU-1 and CIDEr scores compared to the DenseNet201, which has the highest depth in considered CNNs. These results show that accuracies obtained by most of the CNNs are, again, similar to each other.
\begin{table*}[t]
\renewcommand{\arraystretch}{0.1}
\centering
\caption{Results Obtained by the Proposed SD-RSIC When Using Different Embedding Sizes}
\label{table:embed_size}
\begin{tabular}{lcccccccc}
\cmidrule[.8pt]{1-9}
Dataset & Embedding Size ($W$) & {\centering BLEU-1}
& {\centering BLEU-2}
& {\centering BLEU-3}
& {\centering BLEU-4}
& {\centering METEOR}
& {\centering ROUGE-L}
& {\centering CIDEr}
\tabularnewline
\cmidrule[.8pt]{1-9}
\multirow{40}{*}{Sydney-Captions} & 128 & \percentage{0.721} & \percentage{0.615} & \percentage{0.521} & \percentage{0.429} & \percentage{0.324} & \percentage{0.619} & \percentage{1.287}
\tabularnewline \cmidrule{2-9}
& 256 & \percentage{0.730} & \percentage{0.627} & \percentage{0.540} & \percentage{0.456} & \percentage{0.335} & \percentage{0.627} & \percentage{1.408}
\tabularnewline \cmidrule{2-9}
& 512 & \percentageBest{0.761} & \percentageBest{0.666} & \percentageBest{0.586} & \percentageBest{0.517} & \percentageBest{0.366} & \percentageBest{0.657} & \percentageBest{1.690}
\tabularnewline \cmidrule{2-9}
& 1024 & \percentage{0.709} & \percentage{0.595} & \percentage{0.506} & \percentage{0.429} & \percentage{0.351} & \percentage{0.631} & \percentage{1.267}
\tabularnewline
\cmidrule[.8pt]{1-9}
\multirow{40}{*}{UCM-Captions} & 128 & \percentage{0.716} & \percentage{0.631} & \percentage{0.560} & \percentage{0.500} & \percentage{0.362} & \percentage{0.660} & \percentage{1.999}
\tabularnewline \cmidrule{2-9}
& 256 & \percentage{0.742} & \percentage{0.657} & \percentage{0.587} & \percentage{0.523} & \percentage{0.381} & \percentage{0.684} & \percentage{2.028}
\tabularnewline \cmidrule{2-9}
& 512 & \percentageBest{0.748} & \percentageBest{0.664} & \percentageBest{0.598} & \percentageBest{0.538} & \percentage{0.390} & \percentageBest{0.695} & \percentage{2.132}
\tabularnewline \cmidrule{2-9}
& 1024 & \percentage{0.746} & \percentage{0.662} & \percentage{0.594} & \percentage{0.537} & \percentageBest{0.392} & \percentage{0.691} & \percentageBest{2.139}
\tabularnewline
\cmidrule[.8pt]{1-9}
\multirow{40}{*}{RSICD} & 128 & \percentage{0.610} & \percentage{0.432} & \percentage{0.331} & \percentage{0.265} & \percentage{0.230} & \percentage{0.494} & \percentage{0.660}
\tabularnewline \cmidrule{2-9}
& 256 & \percentage{0.634} & \percentage{0.458} & \percentage{0.353} & \percentage{0.283} & \percentage{0.244} & \percentage{0.510} & \percentage{0.736}
\tabularnewline \cmidrule{2-9}
& 512 & \percentage{0.644} & \percentageBest{0.474} & \percentageBest{0.369} & \percentageBest{0.300} & \percentage{0.249} & \percentageBest{0.523} & \percentageBest{0.794}
\tabularnewline \cmidrule{2-9}
& 1024 & \percentageBest{0.647} & \percentage{0.468} & \percentage{0.359} & \percentage{0.288} & \percentageBest{0.250} & \percentage{0.515} & \percentage{0.787}
\tabularnewline
\cmidrule[.8pt]{1-9}
\end{tabular}
\end{table*}

\begin{table*}[t]
\renewcommand{\arraystretch}{0.9}
\centering
\caption{Results Obtained by the Proposed SD-RSIC on the Complete Set of Captions and its first step on a Single Caption for Each Image}
\label{table:single_caption}
\begin{tabular}{llccccccc}
\cmidrule[.8pt]{1-9}
Dataset & Method & {\centering BLEU-1}
& {\centering BLEU-2}
& {\centering BLEU-3}
& {\centering BLEU-4}
& {\centering METEOR}
& {\centering ROUGE-L}
& {\centering CIDEr}
\tabularnewline
\cmidrule[.8pt]{1-9}
\multirow{2}{*}{Sydney-Captions} & Step 1 (Single Caption) & \percentage{0.665} & \percentage{0.553} & \percentage{0.470} & \percentage{0.399} & \percentage{0.312} & \percentage{0.606} & \percentage{1.099}
\tabularnewline \cmidrule{2-9}
& SD-RSIC & \percentageBest{0.761} & \percentageBest{0.666} & \percentageBest{0.586} & \percentageBest{0.517} & \percentageBest{0.366} & \percentageBest{0.657} & \percentageBest{1.690}
\tabularnewline
\cmidrule[.8pt]{1-9}
\multirow{2}{*}{UCM-Captions} & Step 1 (Single Caption) & \percentage{0.700} & \percentage{0.602} & \percentage{0.526} & \percentage{0.460} & \percentage{0.332} & \percentage{0.636} & \percentage{1.774}
\tabularnewline \cmidrule{2-9}
& SD-RSIC & \percentageBest{0.748} & \percentageBest{0.664} & \percentageBest{0.598} & \percentageBest{0.538} & \percentageBest{0.390} & \percentageBest{0.695} & \percentageBest{2.132}
\tabularnewline
\cmidrule[.8pt]{1-9}
\multirow{2}{*}{RSICD} & Step 1 (Single Caption) & \percentage{0.629} & \percentage{0.455} & \percentage{0.352} & \percentage{0.285} & \percentage{0.244} & \percentage{0.510} & \percentage{0.740}
\tabularnewline \cmidrule{2-9}
& SD-RSIC & \percentageBest{0.644} & \percentageBest{0.474} & \percentageBest{0.369} & \percentageBest{0.300} & \percentageBest{0.249} & \percentageBest{0.523} & \percentageBest{0.794}
\tabularnewline
\cmidrule[.8pt]{1-9}
\end{tabular}
\end{table*}

The sensitivity analysis for different CNN models used in the first step shows that utilizing different models does not significantly affect the RS image captioning performance of our approach. However, the proper selection of a CNN model and its depth can improve the performance of the SD-RSIC. Accordingly, we utilized the ResNet101, VGG16 and ResNet152 for the rest of the experiments on Sydney-Captions, UCM-Captions and RSICD datasets, respectively.

In the second set of trials, we assessed the effect of the embedding size $W$ used in the proposed approach. Table~\ref{table:embed_size} shows the image captioning performances under the different sizes of embedding space for all the considered datasets. By assessing the table, one can observe that increasing the value of $W$ up to some extent provides significantly higher scores under all the metrics compared to those obtained by using the lowest value of $W$. As an example, the proposed approach with the $W=512$ provides more than $6\%$ higher BLEU-3, more than $4\%$ higher METEOR and almost $4\%$ higher ROUGE-L scores compared to that of the $W=128$ for the Sydney-Captions dataset. This is due to the fact that increasing the value of $W$ allows to preserve more detailed information than the lowest dimensional embeddings for both image features and image captions. However, selecting a very high value of $W$ (e.g., $W=1024$) does not further improve the information preserving capability of the proposed SD-RSIC. As an example, the proposed approach with the $W=512$ leads to almost $1\%$ higher BLEU-2, BLEU-3, BLUE-4, ROUGE-L, CIDEr scores and almost the same BLEU-1 and METEOR scores compared to that of the $W=1024$ for the RSICD dataset. It is worth noting that increasing the value of $W$ also increases the computational complexity of the proposed approach. Accordingly, we selected the value of $W$ as 512 for the rest of the experiments on the considered datasets.
\begin{figure}[t]
  \centering
  \includegraphics[width=1\linewidth]{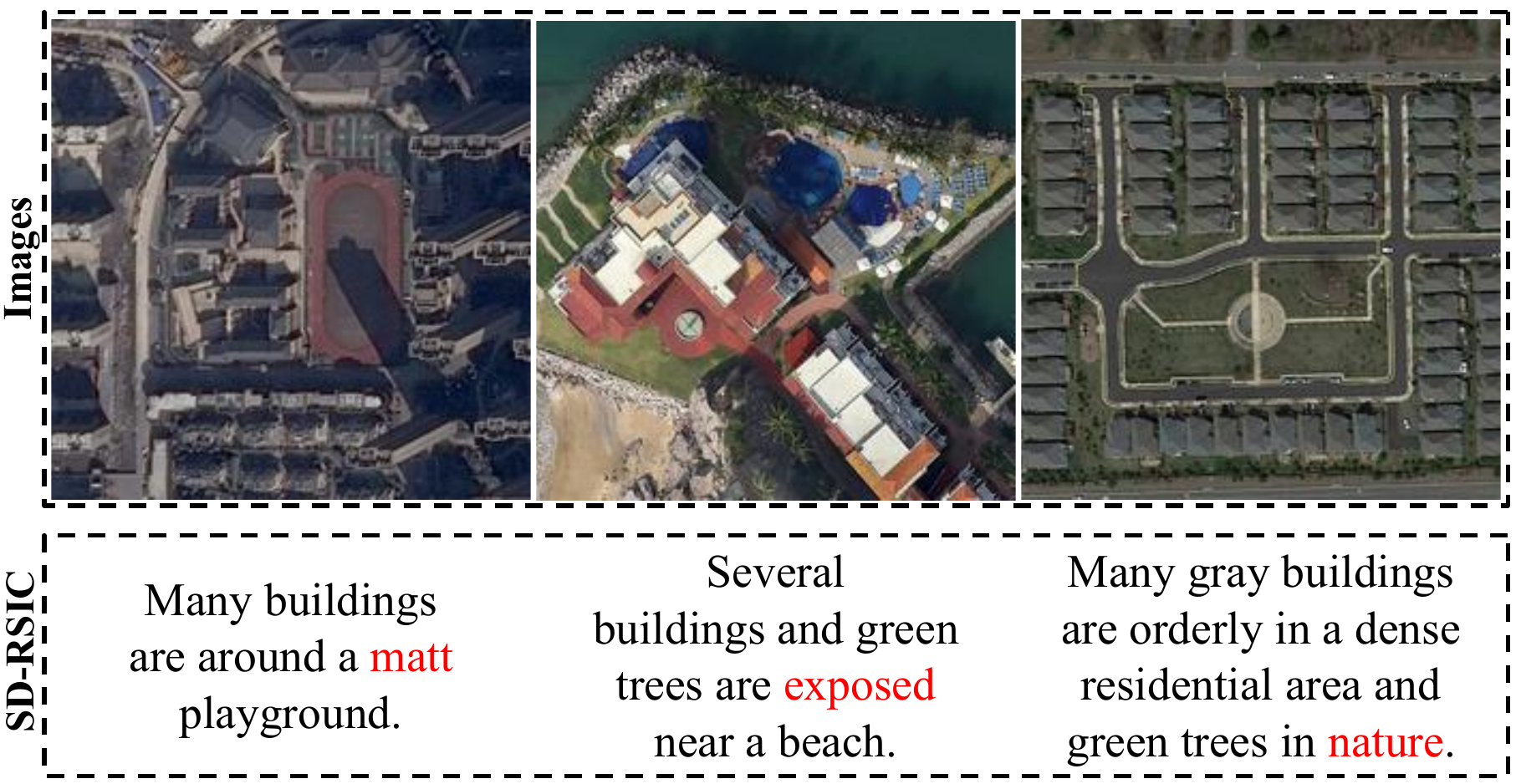}
  \caption{An example of the RSICD images with the generated captions by the SD-RSIC. The words, which are only from Annotated Gigaword dataset, are in red.}
  \label{fig:vis_result2}
\end{figure}

In the third set of trials, we analyzed the effectiveness of the second and third steps of the proposed approach. Table~\ref{table:single_caption} shows the image captioning performances obtained when: i) the first step of the proposed approach is applied by considering only a single caption for each image (i.e., Step 1 (Single Caption)); and ii) the proposed SD-RSIC approach is applied by considering the complete set of captions for all the considered datasets. By analyzing the table, one can see that our proposed approach results in significantly better performances with respect to the Step 1 (Single Caption) for all datasets. As an example, the proposed approach provides almost $6\%$ higher BLUE-1 and more than $5\%$ higher ROUGE-L scores for the Sydney-Captions dataset, while providing more than $7\%$ higher BLEU-3 and BLEU-4 scores for the UCM-Captions dataset compared to the Step 1 (Single Caption). This is due to the fact that the second and the third steps of the SD-RSIC significantly addresses the problems related to redundancy present in ground-truth captions, and thus improves the image captioning performances. Fig.~\ref{fig:vis_result2} shows an example of RSICD images with the generated captions by the SD-RSIC. By assessing the figure, one can observe that the SD-RSIC provides the enriched vocabulary compared to the original vocabulary of captioning datasets. As an example, the words for describing the objects on the ground (e.g., matt, nature) are from the Annotated Gigaword dataset and not included in the original vocabulary of the RSICD dataset. The enriched vocabulary of the proposed approach leads to more detailed captions for semantically complex RS images. These results show that the SD-RSIC overcomes the limitations of redundant information in ground-truth captions, while providing the detailed language semantics due to its second and third steps.

\begin{table}[t] 
\centering
\caption{Number of Required Model Parameters (NP) and Floating-point Operations (FLOPs) Associated to the Different Steps of the Proposed Approach (The Sydney-Captions Dataset)}
\label{table:comp_complexity_Sydney}
\begin{tabular}{
@{\hskip -0.025in}*{3}{>{\centering\arraybackslash}p{0.06\textwidth}}*{2}{>{\centering\arraybackslash}p{0.06\textwidth}}@{\hskip -0.01in}}
\toprule
\multicolumn{3}{c}{Steps of the Proposed Approach} & \multirow{3}{0.06\textwidth}[2pt]{\centering NP\\($\times 10^{6}$)} & \multirow{3}{0.06\textwidth}[2pt]{\centering FLOPs\\($\times 10^{9}$)}
\tabularnewline 
\cmidrule(lr{1em}){1-3}
1\textsuperscript{st}& 2\textsuperscript{nd} & 3\textsuperscript{rd} & & \tabularnewline
\midrule
\cmark & \xmark & \xmark & $43.55$ & $7.83$\tabularnewline
\cmark & \cmark & \xmark & $77.93$ & $8.62$ \tabularnewline
\cmark & \cmark & \cmark & $77.94$ & $8.63$ \tabularnewline
\bottomrule
\end{tabular}
\end{table}

\begin{table}[t]
\centering
\caption{Number of Required Model Parameters (NP) and Floating-point Operations (FLOPs) Associated to the Different Steps of the Proposed Approach (The UCM-Captions Dataset)}
\label{table:comp_complexity_UCM}
\begin{tabular}{
@{\hskip -0.025in}*{3}{>{\centering\arraybackslash}p{0.06\textwidth}}*{2}{>{\centering\arraybackslash}p{0.06\textwidth}}@{\hskip -0.01in}}
\toprule
\multicolumn{3}{c}{Steps of the Proposed Approach} & \multirow{3}{0.06\textwidth}[2pt]{\centering NP\\($\times 10^{6}$)} & \multirow{3}{0.06\textwidth}[2pt]{\centering FLOPs\\($\times 10^{9}$)}
\tabularnewline 
\cmidrule(lr{1em}){1-3}
1\textsuperscript{st}& 2\textsuperscript{nd} & 3\textsuperscript{rd} & & \tabularnewline
\midrule
\cmark & \xmark & \xmark & $119.57$ & $15.46$\tabularnewline
\cmark & \cmark & \xmark & $153.96$ & $16.26$ \tabularnewline
\cmark & \cmark & \cmark & $153.97$ & $16.27$ \tabularnewline
\bottomrule
\end{tabular}
\end{table}

\begin{table}[t]
\centering
\caption{Number of Required Model Parameters (NP) and Floating-point Operations (FLOPs) Associated to the Different Steps of the Proposed Approach (The RSICD Dataset)}
\label{table:comp_complexity_RSICD}
\begin{tabular}{
@{\hskip -0.025in}*{3}{>{\centering\arraybackslash}p{0.06\textwidth}}*{2}{>{\centering\arraybackslash}p{0.06\textwidth}}@{\hskip -0.01in}}
\toprule
\multicolumn{3}{c}{Steps of the Proposed Approach} & \multirow{3}{0.06\textwidth}[2pt]{\centering NP\\($\times 10^{6}$)} & \multirow{3}{0.06\textwidth}[2pt]{\centering FLOPs\\($\times 10^{9}$)}
\tabularnewline 
\cmidrule(lr{1em}){1-3}
1\textsuperscript{st}& 2\textsuperscript{nd} & 3\textsuperscript{rd} & & \tabularnewline
\midrule
\cmark & \xmark & \xmark & $59.18$ & $11.55$\tabularnewline
\cmark & \cmark & \xmark & $93.57$ & $12.35$ \tabularnewline
\cmark & \cmark & \cmark & $93.58$ & $12.35$ \tabularnewline
\bottomrule
\end{tabular}
\end{table}
\begin{table}[t]
\renewcommand{\arraystretch}{1}
\centering
\caption{The Average Rate of Zero-Padding Operation Applied in the Third Step of the Proposed SD-RSIC}
\label{table:zero_padding}
\begin{tabular}{@{\hskip -0.025in}*{3}{>{\centering\arraybackslash}p{0.12\textwidth}}}
\cmidrule[.8pt]{1-3}
Sydney-Captions & UCM-Captions & RSICD
\tabularnewline
\cmidrule[.8pt]{1-3}
 \percentage{0.4245}\% & \percentage{0.324795567}\% & \percentage{0.338541667}\% \tabularnewline
\cmidrule[.8pt]{1-3}
\end{tabular}
\end{table}

In the fourth set of trials, we assessed the computational complexity associated to the different steps of the proposed approach. Table~\ref{table:comp_complexity_Sydney}, \ref{table:comp_complexity_UCM} and \ref{table:comp_complexity_RSICD} show the number of model parameters and the floating-point operations (FLOPs) for the Sydney-Captions, the UCM-Captions and the RSICD datasets, respectively. By analyzing the tables, one can observe that the selection of a CNN model for the first step at the proposed approach is one of the most important factors affecting the overall computational complexity. As an example, the total number of FLOPs for the UCM-Captions dataset is twice as large as that of the Sydney-Captions dataset due to the different CNNs used for these datasets. It is worth noting that this can affect almost all deep learning based image captioning approaches. In addition, the third step of the proposed approach does not significantly affect the computational complexity. As an example, when the third step is included within the proposed approach, the amount of increase in the total number of parameters is less than $1\%$. In greater details, the amount of increase in the FLOPs is significantly less than that in the number of parameters when the second step is included within the proposed approach. These results show that the second step of the proposed approach does not significantly increase the computational time during training. 

In the fifth set of trials, we analyzed the effect of zero-padding operation applied to the summarized captions in the third step of the proposed approach. Table~\ref{table:zero_padding} shows the average rate of zero-padding operation during training for the considered datasets. By assessing the table, one can observe that the zero-padding operation is not often applied to the summarized captions. As an example, for the RSICD dataset, it is applied once in three times on average. To this end, integration of summarized captions with standard captions is not dominated by standard captions. If zero-padding operation is applied frequently, final caption generation may mostly relies on the standard captions. This condition can be eliminated by: 1) using other summarization approaches, which are capable of producing longer sentences, in the second step of the proposed approach; or 2) changing the pre-training of the pointer-generator model (which is utilized in the second step) with different datasets to produce longer sentences.

\begin{table*}[t]
\renewcommand{\arraystretch}{0.9}
\centering
\caption{Results Obtained by the BoW+CNN, DeViSE+CNN, CCSMLF, NIC and the Proposed SD-RSIC (The Sydney-Captions Dataset)}
\label{table:comp_sydney}
\begin{tabular}{lccccccc}
\cmidrule[.8pt]{1-8}
Method & {\centering BLEU-1}
& {\centering BLEU-2}
& {\centering BLEU-3}
& {\centering BLEU-4}
& {\centering METEOR}
& {\centering ROUGE-L}
& {\centering CIDEr}
\tabularnewline
\cmidrule[.8pt]{1-8}
BoW+CNN~\cite{SemanticDescriptions} & \percentage{0.6225}  & \percentage{0.4787}  & \percentage{0.3902}  & \percentage{0.3292}  & \percentage{0.2449}  & \percentage{0.5174}  & \percentage{1.2834}
\tabularnewline \midrule
DeViSE+CNN~\cite{SemanticDescriptions} & \percentage{0.6415}  & \percentage{0.5153}  & \percentage{0.4347}  & \percentage{0.3808}  & \percentage{0.2704}  & \percentage{0.5663}  & \percentage{1.3916}
\tabularnewline \midrule
CSMLF~\cite{SemanticDescriptions} & \percentage{0.4441}  & \percentage{0.3369}  & \percentage{0.2815}  & \percentage{0.2408}  & \percentage{0.1576}  & \percentage{0.4018}  & \percentage{0.9378}
\tabularnewline \midrule
NIC~\cite{GoogleNIC} & \percentage{0.707} & \percentage{0.591} & \percentage{0.503} & \percentage{0.425} & \percentage{0.320} & \percentage{0.606} & \percentage{1.277}
\tabularnewline \midrule
SD-RSIC & \percentageBest{0.761} & \percentageBest{0.666} & \percentageBest{0.586} & \percentageBest{0.517} & \percentageBest{0.366} & \percentageBest{0.657} & \percentageBest{1.690}
\tabularnewline
\cmidrule[.8pt]{1-8}
\end{tabular}
\end{table*}

\begin{table*}[t]
\renewcommand{\arraystretch}{0.9}
\centering
\caption{Results Obtained by the BoW+CNN, DeViSE+CNN, CCSMLF, NIC and the Proposed SD-RSIC (The UCM-Captions Dataset)}
\label{table:comp_ucm}
\begin{tabular}{lccccccc}
\cmidrule[.8pt]{1-8}
Method &  {\centering BLEU-1}
& {\centering BLEU-2}
& {\centering BLEU-3}
& {\centering BLEU-4}
& {\centering METEOR}
& {\centering ROUGE-L}
& {\centering CIDEr}
\tabularnewline
\cmidrule[.8pt]{1-8}
BoW+CNN~\cite{SemanticDescriptions} &  \percentage{0.4059}  & \percentage{0.2547}  & \percentage{0.1842}  & \percentage{0.1435}  & \percentage{0.1439}  & \percentage{0.3659}  & \percentage{0.4159}
\tabularnewline \midrule
DeViSE+CNN~\cite{SemanticDescriptions} &  \percentage{0.3697}  & \percentage{0.1740}  & \percentage{0.0977}  & \percentage{0.0598}  & \percentage{0.0977}  & \percentage{0.2973}  & \percentage{0.0970}
\tabularnewline \midrule
CSMLF~\cite{SemanticDescriptions} & \percentage{0.3874}  & \percentage{0.2145}  & \percentage{0.1253}  & \percentage{0.0915}  & \percentage{0.0954}  & \percentage{0.3599}  & \percentage{0.3703}
\tabularnewline \midrule
NIC~\cite{GoogleNIC} & \percentage{0.726} & \percentage{0.641} & \percentage{0.575} & \percentage{0.517} & \percentage{0.374} & \percentage{0.673} & \percentage{2.006}
\tabularnewline \midrule
SD-RSIC & \percentageBest{0.748} & \percentageBest{0.664} & \percentageBest{0.598} & \percentageBest{0.538} & \percentageBest{0.390} & \percentageBest{0.695} & \percentageBest{2.132}
\tabularnewline
\cmidrule[.8pt]{1-8}
\end{tabular}
\end{table*}

\begin{table*}[t]
\renewcommand{\arraystretch}{0.9}
\centering
\caption{Results Obtained by the BoW+CNN, DeViSE+CNN, CCSMLF, NIC and the Proposed SD-RSIC (The RSICD Dataset)}
\label{table:comp_RSICD}
\begin{tabular}{lccccccc}
\cmidrule[.8pt]{1-8}
Method & {\centering BLEU-1}
& {\centering BLEU-2}
& {\centering BLEU-3}
& {\centering BLEU-4}
& {\centering METEOR}
& {\centering ROUGE-L}
& {\centering CIDEr}
\tabularnewline
\cmidrule[.8pt]{1-8}
BoW+CNN~\cite{SemanticDescriptions} & \percentage{0.2968}  & \percentage{0.1128}  & \percentage{0.0581}  & \percentage{0.0339}  & \percentage{0.0961}  & \percentage{0.2509}  & \percentage{0.1289}  
\tabularnewline \midrule
DeViSE+CNN~\cite{SemanticDescriptions} & \percentage{0.3068}  & \percentage{0.1138}  & \percentage{0.0558}  & \percentage{0.0307}  & \percentage{0.0973}  & \percentage{0.2563}  & \percentage{0.1244}
\tabularnewline \midrule
CSMLF~\cite{SemanticDescriptions} & \percentage{0.5759}  & \percentage{0.3859}  & \percentage{0.2832}  & \percentage{0.2217}  & \percentage{0.2128}  & \percentage{0.4455}  & \percentage{0.5297}
\tabularnewline \midrule
NIC~\cite{GoogleNIC} & \percentage{0.629} & \percentage{0.460} & \percentage{0.358} & \percentage{0.291} & \percentage{0.243} & \percentage{0.515} & \percentage{0.760}
\tabularnewline \midrule
SD-RSIC & \percentageBest{0.644} & \percentageBest{0.474} & \percentageBest{0.369} & \percentageBest{0.300} & \percentageBest{0.249} & \percentageBest{0.523} & \percentageBest{0.794}
\tabularnewline
\cmidrule[.8pt]{1-8}
\end{tabular}
\end{table*}
\subsection{Comparison of the Proposed Approach with the State-of-the-Art Approaches}
In the sixth set of trials, we assessed the effectiveness of the proposed SD-RSIC approach compared to the state-of-the art RS image captioning approaches, which are: the BoW+CNN~\cite{SemanticDescriptions}, the DeViSE+CNN~\cite{SemanticDescriptions}, the CCSMLF~\cite{SemanticDescriptions} and the NIC~\cite{GoogleNIC}. Table~\ref{table:comp_sydney}, \ref{table:comp_ucm} and \ref{table:comp_RSICD} show the corresponding image captioning performances on the Sydney-Captions, UCM-Captions and RSICD datasets, respectively. By analyzing the tables, one can observe that the proposed SD-RSIC approach leads to the highest scores under all metrics for all datasets. As an example, the SD-RSIC outperforms the CSMLF by almost $32\%$ in BLEU-1 and more than $30\%$ in BLEU-3 for the Sydney-Captions dataset, almost $45\%$ in BLEU-2 and more than $44\%$ in BLEU-4 for the UCM-Captions dataset, and almost $8\%$ ROUGE-L and more than $26\%$ in CIDEr for the RSICD dataset. Similar behaviors are also observed while comparing the BoW+CNN and DeViSE+CNN with our approach under different metrics. This shows that modeling image captions based on the joint characterization of language and RS image semantics significantly improves the RS image captioning performance compared to separately describing their semantics and applying matching. In addition, the proposed SD-RSIC approach outperforms the well-known automatic image captioning approach (the NIC) by almost $6\%$ in BLEU-1, more than $9\%$ in BLEU-4 and more than $5\%$ in ROUGE-L for the Sydney-Captions dataset, more than $2\%$ in BLEU-2 and BLUE-3 for the UCM-Captions dataset, and more than $3\%$ in CIDEr and almost $2\%$ in BLEU-1 for the RSICD dataset. This is due to the second and the third steps of the SD-RSIC that integrate the summarization of ground-truth image captions into the widely used CNN and LSTM based encoder-decoder strategy. This shows that the SD-RSIC is capable of: i) eliminating the redundant information in the training set; ii) increasing the generalization capability of the whole neural network; and iii) improving the vocabulary of training sets compared to the existing approaches.

\begin{figure*}[t]
  \centering
  \includegraphics[width=1\linewidth]{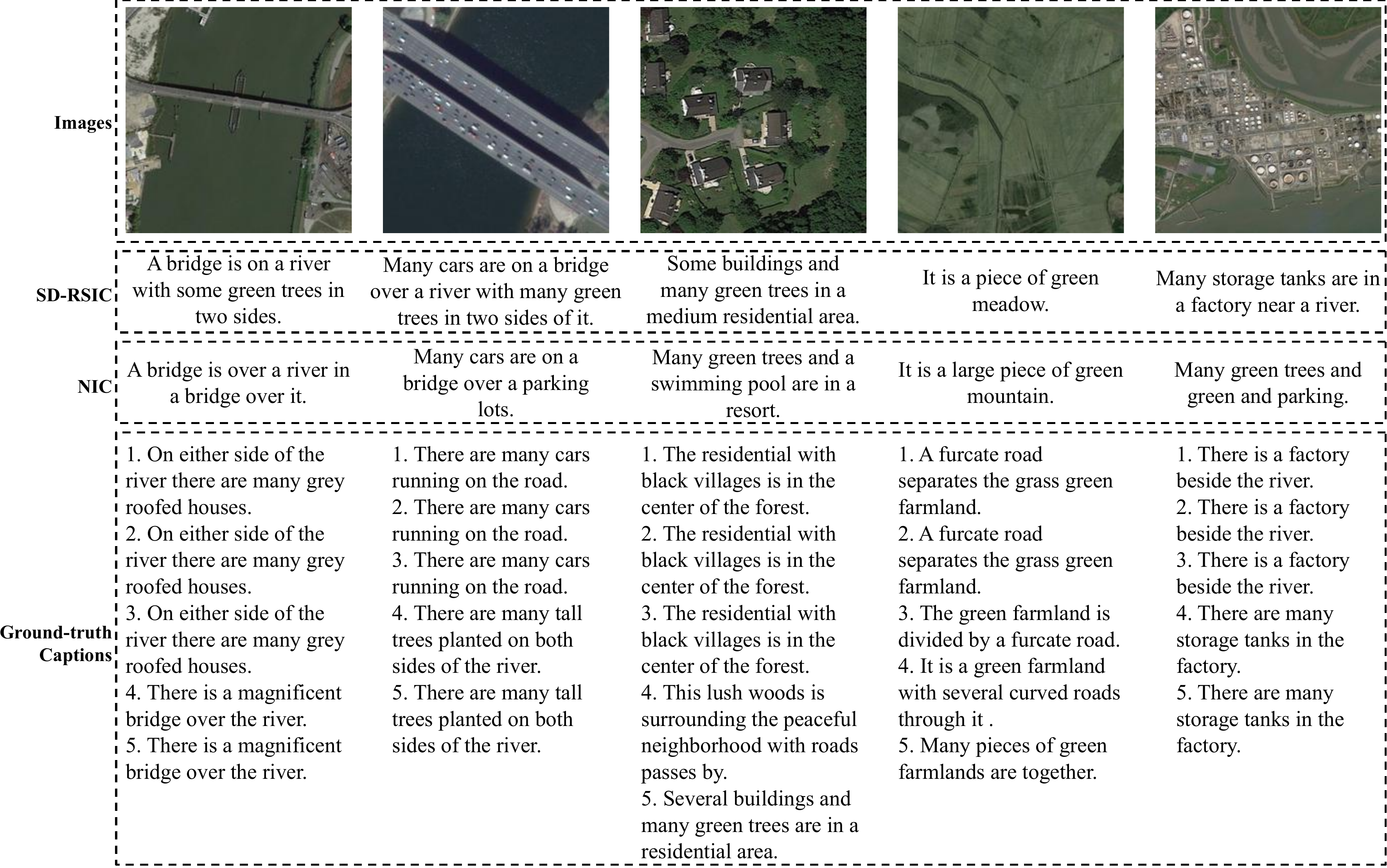}
  \caption{An example of the RSICD images with their five ground-truth captions and the generated captions by the NIC and the SD-RSIC.}
  \label{fig:vis_result}
\end{figure*}
Fig. \ref{fig:vis_result} shows an example of RSICD images with their ground-truth captions and the generated captions by the NIC and the SD-RSIC. By assessing the figure, one can observe that the SD-RSIC provides more accurate image captions to describe the complex semantic content of RS images in the grammatically correct form compared to the NIC. As an example, in the first image, the SD-RSIC is able to describe the green trees near the bridge while this information is not captured by the NIC. In addition to the first image, the SD-RSIC is capable of describing the type of the residential area in the third image that is not characterized in the caption of the NIC. In greater details, for the first and last images, the SD-RSIC is capable of generating the single caption, which accurately describes most of the information associated with the semantic content of the image in a grammatically correct form. However, the NIC provides grammatically incorrect sentences, wrong information in the captions and phrases instead of sentences for the same images. This shows that the SD-RSIC can accurately describe the complex semantic content of RS images with single grammatically correct caption. We observed the similar behaviours for the other approaches and datasets. Thus, qualitative results further confirm that the proposed SD-RSIC approach achieves promising RS image captioning performance.

\section{Conclusion}\label{sec:conclusion}

In this paper, we have introduced a novel Summarization Driven Remote Sensing Image Captioning (SD-RSIC) approach. The proposed SD-RSIC approach consists of three main steps. The first step generates the standard RS image captions by jointly exploiting CNNs and LSTMs. The second step summarizes all ground-truth captions into a single caption by using a sequence-to-sequence deep learning model. Third step automatically computes the adaptive weights for combining the standard captions with summarized captions, relying on the semantic content of RS images based on their image level features. Experimental results obtained on the existing RS image captioning datasets show the effectiveness of the proposed SD-RSIC approach over the state-of-the-art approaches. The main reasons for the success of our proposed SD-RSIC approach are summarized as follows:
\begin{enumerate}
    \item Due to the summarization of ground-truth captions in the second step, the SD-RSIC eliminates the redundant information (occured because of the repetitive as well as highly similar captions) present in the RS image captioning datasets.
    \item Due to the use of the summarization model, which is trained on large text corpora in the second step, the SD-RSIC significantly enriches the image captioning vocabulary in terms of the number and variety of words, resulting in more accurate image captions for complex scenarios. 
    \item Due to the adaptive weights among the standard and summarized captions provided in the third step, which allows effective integration of the condensed (summarized) information of ground-truth captions with standard captions, the SD-RSIC reduces the risk of over-fitting during training and increases the generalization capability of the proposed DNN.
\end{enumerate}

It is worth noting that an attention strategy that finds the most informative regions of RS images in terms of both the generation of standard captions and the integration of summarized captions can further improve the performance of the proposed approach. To this end, any attention strategy presented in the literature can be directly integrated within the proposed approach. We would like to point out that the existing image captioning metrics evaluate the accuracy of the automatically generated image captions by computing the word similarities of these captions with those of the ground truth captions (generated by human experts). These metrics do not compare the actual meaning of the generated and ground truth captions. As a future development of this work we plan to study on defining a new image captioning metric that can intrinsically address this issue. In addition, we also plan to improve the second step of the SD-RSIC by including different: i) summarization approaches (e.g.,~\cite{Yao:2020}); and ii) summarization datasets (e.g., the DUC 2004).

\ifCLASSOPTIONcaptionsoff
  \newpage
\fi



%

\bibliographystyle{IEEEtran}

\bibliography{ref/defs.bib,ref/references.bib}

\begin{thebibliography}{10}
\providecommand{\url}[1]{#1}
\csname url@samestyle\endcsname
\providecommand{\newblock}{\relax}
\providecommand{\bibinfo}[2]{#2}
\providecommand{\BIBentrySTDinterwordspacing}{\spaceskip=0pt\relax}
\providecommand{\BIBentryALTinterwordstretchfactor}{4}
\providecommand{\BIBentryALTinterwordspacing}{\spaceskip=\fontdimen2\font plus
\BIBentryALTinterwordstretchfactor\fontdimen3\font minus
  \fontdimen4\font\relax}
\providecommand{\BIBforeignlanguage}[2]{{%
\expandafter\ifx\csname l@#1\endcsname\relax
\typeout{** WARNING: IEEEtran.bst: No hyphenation pattern has been}%
\typeout{** loaded for the language `#1'. Using the pattern for}%
\typeout{** the default language instead.}%
\else
\language=\csname l@#1\endcsname
\fi
#2}}
\providecommand{\BIBdecl}{\relax}
\BIBdecl

\bibitem{Hoxha:2020}
G.~{Hoxha}, F.~{Melgani}, and B.~{Demir}, ``Toward remote sensing image
  retrieval under a deep image captioning perspective,'' \emph{IEEE J. Sel.
  Top. Appl. Earth Obs. Remote Sens.}, vol.~13, pp. 4462--4475, 2020.

\bibitem{UCMCaptions}
B.~Qu, X.~Li, D.~Tao, and X.~Lu, ``Deep semantic understanding of high
  resolution remote sensing image,'' in \emph{Proc. Intl. Conf. Comput. Inf.
  Telecommunication Syst.}, Jul. 2016, pp. 1--5.

\bibitem{traditionalrscap}
Z.~Shi and Z.~Zou, ``Can a machine generate humanlike language descriptions for
  a remote sensing image?'' \emph{IEEE Trans. Geosci. Remote Sens.}, vol.~55,
  no.~6, pp. 3623--3634, Jun. 2017.

\bibitem{lu2017exploring}
X.~Lu, B.~Wang, X.~Zheng, and X.~Li, ``Exploring models and data for remote
  sensing image caption generation,'' \emph{IEEE Trans. Geosci. Remote Sens.},
  vol.~56, no.~4, pp. 2183--2195, Dec. 2017.

\bibitem{SemanticDescriptions}
B.~Wang, X.~Lu, X.~Zheng, and X.~Li, ``Semantic descriptions of high-resolution
  remote sensing images,'' \emph{IEEE Geosci. Remote Sens. Lett.}, vol.~16,
  no.~8, pp. 1--5, Aug. 2019.

\bibitem{RSAttrAttn}
X.~Zhang, X.~Wang, X.~Tang, H.~Zhou, and C.~Li, ``Description generation for
  remote sensing images using attribute attention mechanism,'' \emph{Remote
  Sens.}, vol.~11, no.~6, Mar. 2019.

\bibitem{GoogleNIC}
O.~Vinyals, A.~Toshev, S.~Bengio, and D.~Erhan, ``Show and tell: A neural image
  caption generator,'' in \emph{Proc. IEEE Conf. Comput. Vis. Pattern Recog.},
  June 2015, pp. 3156--3164.

\bibitem{showattendtell}
K.~Xu, J.~L. Ba, R.~Kiros, K.~Cho, A.~Courville, R.~Salakhutdinov, R.~S. Zemel,
  and Y.~Bengio, ``Show, attend and tell: Neural image caption generation with
  visual attention,'' in \emph{Proc. Intl. Conf. Mach. Learn.}, Jul. 2015, pp.
  2048--2057.

\bibitem{Yu:2019}
N.~{Yu}, X.~{Hu}, B.~{Song}, J.~{Yang}, and J.~{Zhang}, ``Topic-oriented image
  captioning based on order-embedding,'' \emph{IEEE Trans. Image Process.},
  vol.~28, no.~6, pp. 2743--2754, Jun. 2019.

\bibitem{2019:Zakir}
M.~Z. Hossain, F.~Sohel, M.~F. Shiratuddin, and H.~Laga, ``A comprehensive
  survey of deep learning for image captioning,'' \emph{ACM Comput. Surv.},
  vol.~51, no.~6, pp. 118:1--118:36, Feb. 2019.

\bibitem{lstm}
S.~Hochreiter and J.~Schmidhuber, ``Long short-term memory,'' \emph{Neural
  Comput.}, vol.~9, no.~8, pp. 1735--1780, Nov. 1997.

\bibitem{Gers:2000}
F.~A. Gers, J.~Schmidhuber, and F.~Cummins, ``Learning to forget: Continual
  prediction with {LSTM},'' \emph{Neural Comput.}, vol.~12, no.~10, pp.
  2451--2471, Oct. 2000.

\bibitem{PointerGenerator17}
A.~See, P.~Liu, and C.~Manning, ``Get to the point: Summarization with
  pointer-generator networks,'' in \emph{Proc. Annu. Meet. Assoc. Comput.
  Linguistics}, 2017, pp. 1073--1083.

\bibitem{gigaword}
D.~Graff, J.~Kong, K.~Chen, and K.~Maeda, \emph{English gigaword}.\hskip 1em
  plus 0.5em minus 0.4em\relax Linguistic Data Consortium, Philadelphia, 2003.

\bibitem{Napoles}
C.~Napoles, M.~Gormley, and B.~Van, Durme, ``Annotated gigaword,'' in
  \emph{Proc. Joint Workshop Automatic Knowledge Base Construction and
  Web-scale Knowledge Extraction}, Jun. 2012, pp. 95--100.

\bibitem{sydneydataset}
F.~Zhang, B.~Du, and L.~Zhang, ``Saliency-guided unsupervised feature learning
  for scene classification,'' \emph{IEEE Trans. Geosci. Remote Sens.}, vol.~53,
  no.~4, pp. 2175--2184, Apr. 2015.

\bibitem{ucmdataset}
Y.~Yang and S.~Newsam, ``Bag-of-visual-words and spatial extensions for
  land-use classification,'' in \emph{Proc. Intl. Conf. Adv. Geogr. Inf.
  Syst.}, Nov. 2010, pp. 270--279.

\bibitem{Rush:2015}
A.~M. Rush, S.~Chopra, and J.~Weston, ``A neural attention model for
  abstractive sentence summarization,'' in \emph{Proc. Conf. Empirical Methods
  Natural Lang. Process.}, Sep. 2015, pp. 379--389.

\bibitem{Simonyan:2015}
K.~Simonyan and A.~Zisserman, ``Very deep convolutional networks for
  large-scale image recognition,'' in \emph{Proc. Intl. Conf. Learn.
  Represent.}, May 2015, pp. 1--14.

\bibitem{Szegedy:2015}
C.~{Szegedy}, {Wei Liu}, {Yangqing Jia}, P.~{Sermanet}, S.~{Reed},
  D.~{Anguelov}, D.~{Erhan}, V.~{Vanhoucke}, and A.~{Rabinovich}, ``Going
  deeper with convolutions,'' in \emph{Proc. IEEE Conf. Comput. Vis. Pattern
  Recog.}, Jun. 2015, pp. 1--9.

\bibitem{Szegedy:2016}
C.~{Szegedy}, V.~{Vanhoucke}, S.~{Ioffe}, J.~{Shlens}, and Z.~{Wojna},
  ``Rethinking the inception architecture for computer vision,'' in \emph{Proc.
  IEEE Conf. Comput. Vis. Pattern Recog.}, Jun. 2016, pp. 2818--2826.

\bibitem{He:2016}
K.~{He}, X.~{Zhang}, S.~{Ren}, and J.~{Sun}, ``Deep residual learning for image
  recognition,'' in \emph{Proc. IEEE Conf. Comput. Vis. Pattern Recog.}, Jun.
  2016, pp. 770--778.

\bibitem{Huang:2017}
G.~{Huang}, Z.~{Liu}, L.~v.~d. {Maaten}, and K.~Q. {Weinberger}, ``Densely
  connected convolutional networks,'' in \emph{Proc. IEEE Conf. Comput. Vis.
  Pattern Recog.}, Jul. 2017, pp. 2261--2269.

\bibitem{Frome:2013}
A.~Frome, G.~S. Corrado, J.~Shlens, S.~Bengio, J.~Dean, M.~Ranzato, and
  T.~Mikolov, ``{DeViSE}: A deep visual-semantic embedding model,'' in
  \emph{Proc. Adv. Neural Inf. Process. Syst.}, Dec. 2013, pp. 2121--2129.

\bibitem{BLEU}
K.~Papineni, S.~Roukos, T.~Ward, and W.-J. Zhu, ``{BLEU}: A method for
  automatic evaluation of machine translation,'' in \emph{Proc. Annu. Meet.
  Assoc. Comput. Linguistics}, Jul. 2002, pp. 311--318.

\bibitem{Meteor2014}
M.~"Denkowski and A.~Lavie, ``Meteor universal: Language specific translation
  evaluation for any target language,'' in \emph{Proc. Workshop Statistical
  Mach. Translation}, Jun. 2014, pp. 376--380.

\bibitem{Rouge2004}
C.-Y. Lin, ``{ROUGE}: A package for automatic evaluation of summaries,'' in
  \emph{Workshop on Text Summarization Branches Out: Proc. Annu. Meet. Assoc.
  Comput. Linguistics}, Jul. 2004, pp. 74--81.

\bibitem{CIDEr2014}
R.~Vedantam, C.~L. Zitnick, and D.~Parikh, ``{CIDEr}: Consensus-based image
  description evaluation,'' in \emph{Proc. IEEE Conf. Comput. Vis. Pattern
  Recog.}, Jun. 2015, pp. 4566--4575.

\bibitem{Yao:2020}
K.~{Yao}, L.~{Zhang}, D.~{Du}, T.~{Luo}, L.~{Tao}, and Y.~{Wu}, ``Dual encoding
  for abstractive text summarization,'' \emph{IEEE Trans. Cybernetics},
  vol.~50, no.~3, pp. 985--996, 2020.

\end{thebibliography}





%
\vspace{-0.2in}
\begin{IEEEbiography}[{\includegraphics[width=1in,height=1.25in,clip,keepaspectratio]{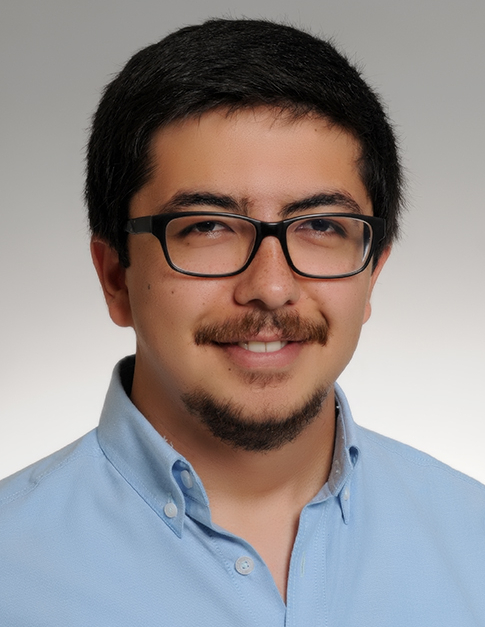}}]{Gencer Sumbul} received his B.S. degree in Computer Engineering from Bilkent University, Ankara, Turkey in 2015 and the M.S. degree in Computer Engineering from Bilkent University in 2018. He is currently a research associate in the Remote Sensing Image Analysis (RSiM) group and pursuing the Ph.D. degree at the Faculty of Electrical Engineering and Computer Science, Technische Universit\"at Berlin, Germany since 2019. His research interests include computer vision, pattern recognition and machine learning, with special interest in deep learning, large-scale image understanding and remote sensing. He is a referee for journals such as the IEEE Transactions on Geoscience and Remote Sensing, the IEEE Access, the IEEE Geoscience and Remote Sensing Letters, the ISPRS Journal of Photogrammetry and Remote Sensing and international conferences such as European Conference on Computer Vision and IEEE International Geoscience and Remote Sensing Symposium.
\end{IEEEbiography}
\vspace{-0.2in}
\begin{IEEEbiography}[{\includegraphics[width=1in,height=1.25in,clip,keepaspectratio]{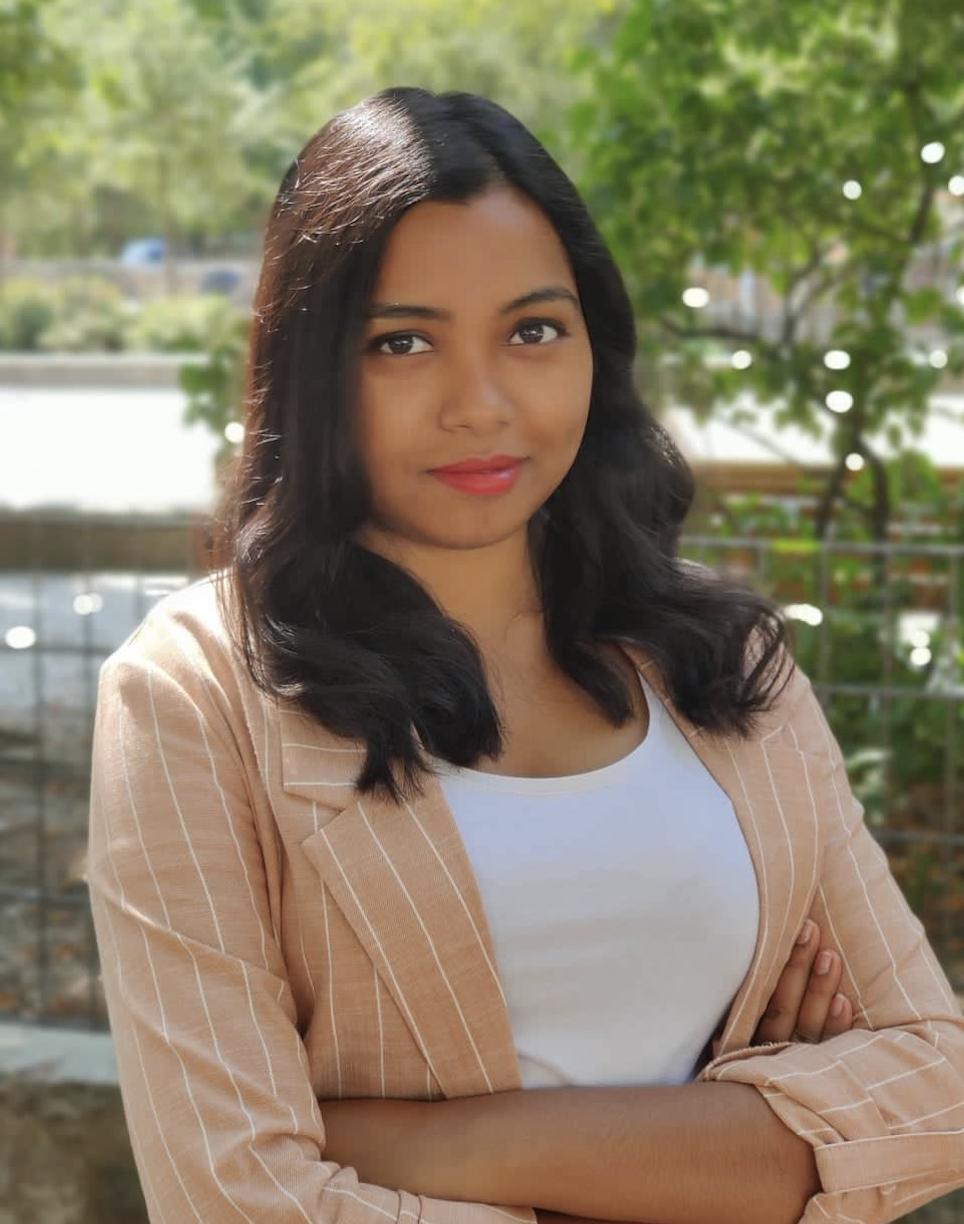}}]{Sonali Nayak}
received the B.S. degree in Computer Science from International Institute of Information Technology Bhubaneswar, India in 2013 and the M.S. degree also in Computer Science from TU Berlin, Germany in 2019. Her research areas include remote sensing image analysis, natural language processing and machine learning.
\end{IEEEbiography}
\vspace{-0.2in}
\begin{IEEEbiography}[{\includegraphics[width=1in,height=1.25in,clip,keepaspectratio]{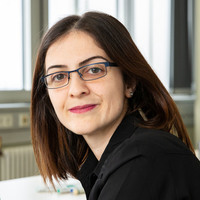}}]{Beg\"{u}m Demir} (S'06-M'11-SM'16) received the B.S., M.Sc., and Ph.D. degrees in electronic and telecommunication engineering from Kocaeli University, Kocaeli, Turkey, in 2005, 2007, and 2010, respectively.

She is currently a Full Professor and head of the Remote Sensing Image Analysis (RSiM) group at the Faculty of Electrical Engineering and Computer Science, Technische Universit\"at Berlin, Germany since 2018. Before starting at TU Berlin, she was an Associate Professor at the Department of Computer Science and Information Engineering, University of Trento, Italy. Her research activities lie at the intersection of machine learning, remote sensing and signal processing. Specifically, she performs research on developing innovative methods for addressing a wide range of scientific problems in the area of remote sensing for Earth observation. She was a recipient of a Starting Grant from the European Research Council (ERC) with the project "BigEarth-Accurate and Scalable Processing of Big Data in Earth Observation" in 2017, and the "2018 Early Career Award" presented by the IEEE Geoscience and Remote Sensing Society. Dr. Demir is a senior member of IEEE since 2016.

Dr. Demir is a Scientific Committee member of several international conferences and workshops, such as: Conference on Content-Based Multimedia Indexing, Conference on Big Data from Space, Living Planet Symposium, International Joint Urban Remote Sensing Event, SPIE International Conference on Signal and Image Processing for Remote Sensing, Machine Learning for Earth Observation Workshop organized within the ECML/PKDD. She is a referee for several journals such as the PROCEEDINGS OF THE IEEE, the IEEE TRANSACTIONS ON GEOSCIENCE AND REMOTE SENSING, the IEEE GEOSCIENCE AND REMOTE SENSING LETTERS, the IEEE TRANSACTIONS ON IMAGE PROCESSING, Pattern Recognition, the IEEE TRANSACTIONS ON CIRCUITS AND SYSTEMS FOR VIDEO TECHNOLOGY, the IEEE JOURNAL OF SELECTED TOPICS IN SIGNAL PROCESSING, the International Journal of Remote Sensing), and several international conferences. Currently she is an Associate Editor for the IEEE GEOSCIENCE AND REMOTE SENSING LETTERS and MDPI Remote Sensing.
\end{IEEEbiography}
\vfill


\end{document}